\documentclass{article} %
\usepackage{collas2026_conference,times}
\usepackage{easyReview}
\usepackage{algorithm}
\usepackage{algorithmic}
\usepackage{booktabs}
\usepackage{multirow}
\usepackage{threeparttable}
\usepackage{wrapfig}

\usepackage{amsmath,amsfonts,bm}

\def\eqref#1{equation~\ref{#1}}

\def\1{\bm{1}}

\def\vtheta{{\bm{\theta}}}

\def\vp{{\bm{p}}}

\def\vu{{\bm{u}}}

\def\vx{{\bm{x}}}

\def\mA{{\bm{A}}}

\def\mF{{\bm{F}}}

\def\mH{{\bm{H}}}
\def\mI{{\bm{I}}}

\def\mQ{{\bm{Q}}}
\def\mR{{\bm{R}}}

\def\mW{{\bm{W}}}

\def\mLambda{{\bm{\Lambda}}}

\DeclareMathAlphabet{\mathsfit}{\encodingdefault}{\sfdefault}{m}{sl}
\SetMathAlphabet{\mathsfit}{bold}{\encodingdefault}{\sfdefault}{bx}{n}

\usepackage{hyperref}
\hypersetup{
    colorlinks=true,
    linkcolor=black,
    filecolor=magenta,
    urlcolor=blue,
    citecolor=purple,
    pdftitle={NORACL},
    pdfpagemode=FullScreen,
    }
    
\usepackage[capitalize, poorman]{cleveref}
\usepackage[docdef=atom]{glossaries-extra}
    \setabbreviationstyle[acronym]{long-short}
    \glssetcategoryattribute{acronym}{nohyperfirst}{true}

    \newacronym{mlp}{MLP}{Multi-layer Perceptron}
\newacronym{ce}{CE}{Cross-Entropy Loss}
\newacronym{mse}{MSE}{Mean Squared Error}
\newacronym{pc}{PC}{Principal Component}
\newacronym{cl}{CL}{Continual Learning}
\newacronym{ewc}{EWC}{Elastic Weight Consolidation}
\newacronym{si}{SI}{Synaptic Intelligence}
\newacronym{ed}{ED}{Effective Dimension}
\newacronym{sota}{SoTA}{State-of-the-Art}
\newacronym{noracl}{NORACL}{ Neurogenesis for Oracle-free Resource-Adaptive Continual Learning}
\newacronym{sgd}{SGD}{Stochastic Gradient Descent}

\usepackage{placeins}

\newif\ifrebuttal
\rebuttaltrue   %

\ifrebuttal
\else
  
  \renewcommand{\remove}[1]{}

  \renewcommand{\comment}[1]{}
\fi

\title{NORACL: Neurogenesis for Oracle-free Resource-Adaptive Continual Learning}

\author{Karthik Charan Raghunathan, Christian Metzner, Laura Kriener$^{*}$, Melika Payvand$^{*}$  \\
Institute of Neuroinformatics\\
University of Zurich \& ETH Zurich\\
Zurich, Switzerland \\
$^{*}$ Shared senior authorship \\
\texttt{\{karthik,melika\}@ini.uzh.ch} \\
}

\preprintcopy %

\begin{document}

\maketitle

\begin{abstract}
\label{sec:abstract}

In a continual learning (CL) setting, we require a model to be plastic enough to learn a new task, but at the same time stable enough to not disturb previously learned capabilities.
We argue that this dilemma has an architectural root. 
A finite network has limited representational and plastic resources, yet the required capacity depends on two properties of the future task stream that are unknown in advance: how many tasks will be encountered (task count), and how much they overlap in feature space (task geometry).
Regularization-based CL methods preserve past knowledge within fixed-capacity architectures and therefore implicitly rely on an \emph{oracle architecture} sized for this unknown future.
When tasks are only weakly related, fixed architectures progressively run out of plastic resources; when tasks are few or strongly overlapping, models are often over-provisioned.

Inspired by neurogenesis in biology, we propose \textbf{NORACL}, a CL framework that addresses the stability-plasticity dilemma by tackling the oracle architecture problem through on-demand neuronal growth. 
Starting from a compact network, NORACL grows only when needed by monitoring two complementary signals for representational and plasticity saturation.
These signals determine when, where, and how much to grow, while preserving previously-learnt behavior.

We evaluate NORACL against oracle-sized static baselines across varying task counts and task geometries. 
Across all settings, NORACL achieves final average accuracies that are better than or on par with oracle-provisioned static baselines while using 10-20\% fewer parameters. Additionally, NORACL yields architectures with interpretable growth, i.e. dissimilar tasks predominantly expand earlier feature-extraction layers, whereas tasks which rely on common features shift growth toward later feature-combination layers. 
Our analysis further explains why fixed-capacity networks lose plasticity as tasks accumulate, even with regularization, whereas NORACL preserves prior knowledge and creates fresh capacity for new tasks through growth. 
Together, these results show that adaptive neurogenesis pushes the stability-plasticity Pareto frontier of continual learning.
\end{abstract}

\section{Introduction}\label{sec:intro}
\begin{figure*}[t]
    \centering
    \includegraphics[width=0.97\textwidth]{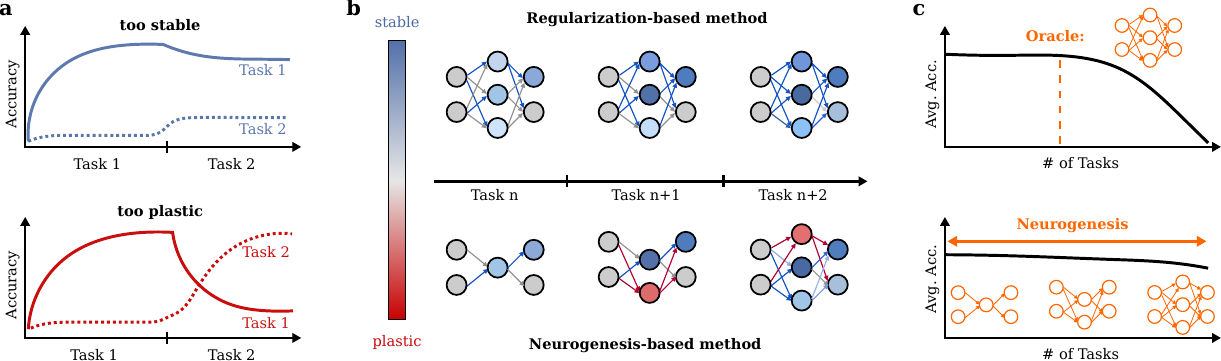}
    \caption{%
    \textbf{The oracle architecture problem.}
    \textbf{a)} ~The stability-plasticity dilemma: a network that is too stable (top) preserves Task~1 but cannot learn Task~2; a network that is too plastic (bottom) learns Task~2 but overwrites Task~1. 
    \textbf{b)} ~Regularization-based methods (top) protect important parameters but progressively exhaust the available plastic capacity of a fixed-size network; neurogenesis-based methods (bottom) add fresh, plastic neurons to absorb new tasks. 
    \textbf{c)} ~Oracle-provisioned static architectures (top) require advance knowledge of the task stream length to set the network size; NORACL (bottom) grows on demand and adapts its capacity to the task stream without such knowledge.
    \label{fig:intro}
    }\vspace{-0.5cm}
\end{figure*}

\Gls{cl} aims to learn new tasks sequentially without forgetting previously learned ones.
Its central challenge is the stability–plasticity dilemma: the network must remain stable enough to preserve past knowledge, yet plastic enough to incorporate new information (\cref{fig:intro}a). 
This trade-off is constrained by the architecture itself, because a finite network has limited representational and plastic resources to allocate across tasks \citep{dohare2024loss, lyle2023understanding}. In \gls{cl}, the required capacity depends on two properties of the future task stream that are unknown in advance: the number of tasks the learner will encounter (task count), and how much structure those tasks will share (task geometry). 
A compact network may suffice for a short sequence of related tasks, whereas more capacity may be needed when tasks are numerous or largely orthogonal. 
Determining the network capacity a~priori requires extensive tuning even for a single task in a non-\gls{cl} setting. 
In \gls{cl}, this difficulty is compounded by the unknown nature of the future task stream.

Regularization-based solutions for \gls{cl} such as \gls{ewc}~\citep{kirkpatrick2017EWC}, Synaptic Intelligence (SI)~\cite{zenke2017continual}, and Memory Aware Synapses (MAS)~\citep{aljundi2018memory} improve stability and mitigate forgetting by protecting parameters that are important for previous tasks, which in turn also reduces the plasticity available for future learning. 
This is effective when future tasks can reuse those parameters, but these methods still operate on fixed-capacity architectures. 
As a result, their success depends implicitly on the architecture having been provisioned for the unknown future task stream, both in the task count and task geometry. 
When tasks are weakly related, there is little opportunity for reuse, regularization progressively freezes the network, and learning stalls because the architecture lacks free resources and plasticity (\cref{fig:intro}b). 
We refer to this hidden dependence of the network architecture on future task count and task overlap as the \emph{oracle architecture problem}.

Biological neural systems face an analogous challenge and address it, in part, through adult neurogenesis. 
For example, in the Hippocampal dentate gyrus, new granule cells are continuously incorporated into existing circuitry throughout life~\citep{aimone2009computational,deng2010new}. 
Computational studies suggest that these newborn neurons support the encoding of novel information while mature neurons preserve established representations~\citep{aimone2009computational,deng2010new,gozel2021functional}. 
This biological strategy suggests that open-ended learning may benefit from adding fresh, highly plastic resources rather than forcing all future learning into a fixed parameter budget (\cref{fig:intro}c).

Inspired by this biological mechanism, we propose \gls{noracl}: a framework that addresses the stability-plasticity dilemma with no \textit{a priori} knowledge about the count and geometry of the future task streams. 
\gls{noracl} starts from a compact initial network and grows when needed by monitoring two complementary signals for the network capacity and plasticity: (i) \gls{ed} of layer activations~\citep{maile2022and} to detect when the network's representational capacity is saturated, and (ii) the cumulative diagonal of the Fisher Information matrix \citep{kirkpatrick2017EWC, schwarz2018progress} to detect when the network's plasticity is exhausted.
When both signals cross their respective thresholds, \gls{noracl} expands the relevant layer by adding fresh neurons in a function-preserving manner, so that new capacity is introduced without disrupting previously acquired knowledge. 
In this way, representational saturation determines when additional capacity is needed, while plasticity saturation confirms that the existing parameters are too constrained to absorb new information. 
Thus, \gls{noracl} only grows when the current architecture is indeed insufficient.
This directly targets the stability–plasticity dilemma: previously learned representations can remain stable, while newly added neurons provide fresh plasticity for future tasks. 
By allowing capacity to expand when needed, especially when future tasks are orthogonal and reuse is limited, \gls{noracl} takes an important step towards \gls{cl} systems that adapt their resources to the uncertainty of the real world. 
As such, \gls{noracl} shifts the accuracy-capacity Pareto frontier of continual learning.

In summary, our contributions are as follows:

    \textbf{(1)} We propose \gls{noracl}, a continual learning method that resolves the oracle architecture problem through on-demand neuronal growth, combining spectral capacity monitoring with plasticity-aware growth triggers to drive efficient neurogenesis.

    \textbf{(2)} %
    We theoretically show on an analytically tractable task stream that, while in the current de-facto \gls{sota} method, \gls{ewc}~\citep{kirkpatrick2017EWC}, the effective plasticity of the network decays monotonically as tasks accumulate, \gls{noracl} shows no such decline due to it's on-demand neuronal growth.

    \textbf{(3)} We show empirically that  \gls{noracl} using no prior knowledge of the future task stream, achieves final accuracies on par with oracle-provisioned static baselines, while using 10-20\% fewer parameters across varying task counts and varying task geometries. 

    \textbf{(4)} We show that \gls{noracl} yields interpretable growth patterns with respect to task geometry: weakly similar tasks predominantly expand earlier feature-extraction layers, whereas more similar tasks shift growth toward later feature-combination layers.

\section{Background \& Related Work} \label{sec:related_work}
\paragraph{Regularization-based methods}
Regularization-based methods promote stability and avoid catastrophic forgetting \citep{mccloskey1989catastrophic, french1999catastrophic} by constraining updates on parameters important for previous tasks.
Notably, \gls{ewc} augments the loss with a quadratic penalty on deviations from previously learned parameters, weighted by the diagonal of the Fisher information matrix as an estimate of parameter importance \citep{kirkpatrick2017EWC}.
Subsequent methods, such as Synaptic Intelligence \citep{zenke2017continual} and Memory-Aware Synapses \citep{aljundi2018memory}, use online or perturbation-based estimates of importance, but share the mechanism of protecting previously important parameters from being overwritten.
While effective at maintaining stability, these regularization-based approaches operate on fixed-capacity architectures and therefore inherently limit plasticity.
As new tasks stream in, an increasing fraction of parameters becomes important for some task and hence constrained, reducing plasticity even in the absence of direct interference between task information.
Consequently, these methods require sufficient capacity to be allocated in advance based on \textit{a priori} knowledge about the task count and task geometry that will be encountered.

\paragraph{Expansion-based methods}

A broad line of continual learning work addresses the fixed-capacity limitation by adapting the network's architecture or its effective capacity over time. These approaches span parameter-isolation and pruning, modular and routing methods, and explicit neuronal growth, and we review each in turn before positioning \gls{noracl}.
Progressive Neural Networks \citep{rusu2016progressive} add task-specific sub-networks for each new task, maintaining plasticity by construction.
However, this requires knowledge of the task identity at inference time such that inputs are correctly routed through the appropriate task networks as well as ignoring the potential advantage of exploiting shared features between tasks.
Dynamically Expandable Networks \citep{yoon2017lifelong} instead grow the network at the level of individual neurons, using group-sparse regularization to identify parameters for selectively retraining at each new task.
Growth is triggered when encountering loss plateaus during this selective retraining phase.
Similar to Progressive Neural Networks, this approach relies on task identity to ensure that neurons added later in the task stream throughout the entire final network are not used for older tasks during inference.
Other expansion-based methods include Reinforced Continual Learning \citep{xu2018reinforced}, which uses a computationally expensive reinforcement learning agent to add or remove neurons, and Compacting, Picking, and Growing \citep{hung2019compacting}, which frees unused capacity through pruning before resorting to expansion.
More recently, Dynamically Expandable Representation \citep{yan2021der} expands capacity by appending new adaptively pruned feature extractors for each new task in a class-incremental learning set-up while maintaining a shared classifier head.
A common limitation across these expansion-based methods is that growth decisions are based on heuristic criteria, such as loss plateaus or sparsity levels, which can conflate optimization difficulty with genuine capacity limitations.
Moreover, both Progressive Neural Networks and Dynamically Expandable Networks require knowledge about the task identity during inference, making them unsuitable for domain-incremental learning.
In addition, because they grow by allocating independent, task-specific capacity (such as DER's sequential feature extractors), their feature sharing is strictly unidirectional. While new tasks can leverage previous representations, they cannot refine the shared feature space to benefit earlier tasks.

\paragraph{Parameter isolation, modularity, and architecture search.} 
A related family keeps total capacity fixed but dynamically allocates it across tasks. Pruning- and mask-based methods such as PackNet \citep{mallya2018packnet}, Piggyback \citep{mallya2018piggyback}, HAT \citep{serra2018overcoming}, and SupSup \citep{wortsman2020supermasks} carve task-specific sub-networks from a shared backbone, while modular and routing methods compose tasks from a library of reusable components, PathNet \citep{fernando2017pathnet} evolves routes through a fixed super-network, and MNTDP \citep{veniat2020efficient} instantiates or reuses modules under a task-driven prior. A further line treats architecture itself as the object of search, using differentiable NAS \citep{li2019learn} to decide, per task, whether to reuse, adapt, or add structure. Most recently, SERENA \citep{yildirim2024self} routes each concept through a dedicated path in an over-parameterized network, freezing it once learned. Across this landscape, three properties recur that distinguish these methods from our setting: they predominantly (i) require task identity at inference to select the correct mask, module, or path; (ii) trigger structural change through heuristic signals (loss or validation plateaus, sparsity levels) or through an auxiliary controller whose cost can exceed that of the task network itself; and (iii) allocate task-specific capacity that is frozen or masked, so feature sharing is unidirectional and earlier tasks cannot benefit from later refinement.

\paragraph{Principled Growth Signals}

Recent work has explored more principled signals for network expansion in the single-task setting.
The NORTH* framework \citep{maile2022and} monitors the \gls{ed} of layer activations to determine when a layer reaches its representational capacity limit, trigger layer expansion when this measure exceeds a threshold defined at network initialization.
\gls{ed} is computed via singular value decomposition over the activations, providing a data-driven and task-identity agnostic signal for growth that does not rely on arbitrary loss or sparsity thresholds nor explicit task information.
However, the application of the \gls{ed} signal to continual learning remains unexplored.
In particular, it is unclear whether using it as a growth trigger would sufficiently sustain plasticity or efficiently exploit shared task geometry across long task streams.
In addition, it fails to fully address the stability-plasticity trade-off, as it lacks mechanisms to protect previously learnt parameters from future interference.
In continual learning, such signals may also reveal how architectural growth adapts to task geometry, an aspect we analyze empirically in this work.

Building on top of this background, \gls{noracl} proposes a principled growth criterion (a parameter-free trigger that jointly gates on spectral \gls{ed} and plasticity signals) without dependence on task identity and additional auxiliary controller. This combination lets \gls{noracl} operate in the strict domain-incremental setting, where explicit task identity information is not provided, while grounding the growth decision in internal capacity signals rather than heuristic observables.

\section{Methods} \label{sec:methods}

We consider a domain-incremental continual learning setting~\citep{van2019three}. 
A model receives a sequence of tasks $\mathcal{D}_1, \mathcal{D}_2, \ldots, \mathcal{D}_T$ with $\mathcal{D}_t = \{(\vx_j^{(t)}, y_j^{(t)})\}_{j=1}^{N_t}$ drawn from task-specific input distributions, but a shared label space $\mathcal{Y}$. At each stage $t$, only the current task's data $\mathcal{D}_t$ is available, no prior data can be revisited, and no task identity is provided at inference.
Crucially, the total number of tasks $T$ is not known in advance.
In this work, we use $L$-layer ReLU \glspl{mlp} with parameters $\vtheta$. 
Hidden layer widths $M_l$ are initialized as $M_l^{(0)}$ and may grow during training, while the output dimension is fixed at $|\mathcal{Y}|$ (the number of classes).
We use plain \gls{sgd} for optimization as it is well motivated for continual learning \citep{mirzadeh2020understanding}.
Final performance is evaluated using average accuracy after training on all $T$ tasks. 

\gls{noracl}'s growth mechanism must answer three questions: \emph{when} to grow (is the current architecture insufficient?), \emph{where} to grow (which layer is the bottleneck?), and \emph{how} to grow (how should new neurons be initialized?). While growth injects fresh plasticity, we still need a mechanism to protect existing representations from interference. We therefore pair the growth mechanism with an online \gls{ewc} \citep{schwarz2018progress} consolidation backbone. After each task $t$, the diagonal of the Fisher information matrix, which estimates how important each parameter is for the tasks seen so far, is computed and accumulated across tasks via exponential moving average:
\begin{equation}
\label{eq:fisher_blend}
\tilde{\mF}^{(t)} = \alpha \, \tilde{\mF}^{(t-1)} + (1-\alpha) \, \mF^{(t)},
\end{equation}
where $\alpha \in (0,1)$ is a blending factor, $\mF^{(t)}$ denotes the Fisher computed on task $t$ alone, while $\tilde{\mF}^{(t)}$ denotes the accumulated running average after task $t$. The training loss for task $t$ ($t \geq 2$) combines the cross-entropy on the current task with a quadratic penalty anchored at the parameters $\vtheta^*$ from the end of the previous task:
\begin{equation}
\label{eq:grohess_oewc_loss}
\mathcal{L}_t = \mathcal{L}_{\mathrm{CE}}(\vtheta; \mathcal{D}_t) + \frac{\lambda}{2} \sum_i \tilde{\mF}_i^{(t-1)} (\theta_i - \theta_i^*)^2.
\end{equation}
where $\theta_i^*$ are the parameters after task $t-1$. 
Task 1 is unregularized (since there is no prior knowledge to protect). \gls{noracl}'s growth mechanism is agnostic to the specific regularization used; we use \gls{ewc} for its simplicity and because it provides the Fisher diagonal needed by the growth trigger below.

\subsection{Growth Trigger: When and Where to Add Neurons}
\label{sec:method_trigger}

\gls{noracl} monitors the network's state at each training epoch and only triggers growth when two signals jointly indicate insufficient reusable capacity: the layer's representational capacity is saturated, and the layer's parameters are too important for previous tasks to be overwritten (plasticity saturation).

\textbf{(1) Representational saturation:}
Given layer $l$ activations $\mH_l \in \mathbb{R}^{M_l \times n}$ over a mini-batch of $n$ samples, the normalized \glsfirst{ed}~\citep{maile2022and} is defined by the set cardinality of $\mH_l$'s singular values:
\begin{equation}
\label{eq:ed}
\varphi_l = \frac{1}{M_l} \left| \left\{ \sigma_i \in \mathrm{SVD}\!\left(\frac{1}{\sqrt{n}} \mH_l\right) \;\middle|\; \sigma_i > \varepsilon \right\} \right|,
\end{equation}
where $\varepsilon > 0$ is a small threshold and SVD stands for singular value decomposition. 
\gls{ed} measures what fraction of the layer's neurons produce linearly independent activation patterns across the current batch. 
A value of $\varphi_l \approx 1$ indicates that all neurons are producing near-orthogonal features and the layer is operating at full representational capacity.

Growth triggers when $\varphi_l > \gamma \cdot \varphi_l^{(0)}$, where $\varphi_l^{(0)}$ is the \gls{ed} measured at the \emph{end of the previous task} (after training) and $\gamma \in (0,1)$ is a sensitivity discount. The discount $\gamma$ is necessary because \gls{ed} fluctuates during training due to batch variability and comparing against the full reference $\varphi_l^{(0)}$ would trigger growth only when the layer is already saturated, leaving no margin to act proactively. Discounting by $\gamma$ allows growth to begin slightly before full saturation is reached. After each task completes, $\varphi_l^{(0)}$ is reset to the current post-consolidation \gls{ed} value, ensuring the reference tracks the network's evolving capacity rather than comparing against a stale initialization-time snapshot. When the trigger fires, the number of added neurons is:
\begin{equation}
\label{eq:k_neurons}
k_l = \left\lfloor M_l \cdot \left(\varphi_l - \gamma \cdot \varphi_l^{(0)}\right) \right\rfloor,
\end{equation}
with $k_l$ rounded up to 1 if $0<k_l<1$. 
Layers under greater representational pressure thus receive more new neurons.

\textbf{(2) Plasticity saturation:}
The \gls{ed} signal alone could fire spuriously due to transient batch effects. To confirm a layer actually should grow, we want to ensure that the layer's plasticity has actually saturated.
In other words, all the parameters are too important to be overwritten for our current task $t$ without damaging performance on previous tasks.

At each epoch during training on task $t$, we compute the diagonal Fisher of the \emph{current task's data} $\mF_l^{(\mathrm{curr})}$ over the first few mini-batches of $\mathcal{D}_t$ (this is a fresh, within-epoch computation, distinct from the accumulated $\tilde{\mF}^{(t)}$ which is updated only at the end of each task via Eq.~\ref{eq:fisher_blend}). 
Growth requires:
\begin{equation}
\label{eq:fisher_gate}
\mathrm{Percentile}\!\left(\mF_l^{(\mathrm{curr})},\; p\right) > \tau_l^{(t)},
\end{equation}
Here $p$ is a fixed percentile and $\tau_l^{(t)}$ is an exponential moving average of past accumulated Fisher magnitudes, updated at the end of each task: $\tau_l^{(t)} = \alpha \, \tau_l^{(t-1)} + (1-\alpha) \, \mathrm{mean}(\tilde{\mF}_l^{(t)})$.
This condition ensures that at least $(100{-}p)\%$ of the layer's parameters have Fisher values exceeding the historical baseline, indicating that most of the layer's existing parameters are genuinely important for past learnt behavior and that new parameters are really necessary.

\textbf{Combined trigger: }Growth at layer $l$ occurs when \emph{both} conditions are met:
\begin{equation}
\label{eq:combined_trigger}
\mathrm{grow}_l = \underbrace{(\varphi_l > \gamma \cdot \varphi_l^{(0)})}_{\text{ED saturated}} \;\wedge\; \underbrace{(\mathrm{Percentile}(\mF_l^{(\mathrm{curr})}, p) > \tau_l)}_{\text{Fisher saturated}}.
\end{equation}
This joint gating ensures that capacity is added only when the layer is both representationally saturated (all neurons producing independent features) and plasticity-wise saturated (parameters are all important for past and current tasks). 
A cool-down period of $C$ epochs follows each growth event, giving the network time to integrate the newly added neurons before growing again.

\subsection{Neurogenesis: How to Add Neurons}
\label{sec:method_growth}

When \gls{noracl} adds $k_l$ neurons to layer $l$, their fan-in weight vectors are initialized from a random orthogonal basis via QR decomposition~\citep{saxe2013exact} and scaled by a fixed factor $s_{\mathrm{init}}$, which is set to $0.2$ in all our experiments. 
The corresponding fan-out weights into layer $l+1$ are initialized to zero so that newly added neurons do not affect downstream computation at the moment of insertion.
The accumulated Fisher diagonal and the \gls{ewc} anchor parameters are padded with zeros for the new dimensions to ensure that new parameters remain fully unconstrained.
After growth, we re-initialize the optimizer state so that it matches the expanded parameter tree.

\subsection{Algorithmic Properties} 
Because the fan-out weights of newly added neurons are initialized to zero, the network function is unchanged at the moment of growth, i.e.\ $f(\vx; \vtheta') = f(\vx; \vtheta)$ for all inputs $\vx$, aiding the \emph{functional stability} of the network on previously learned tasks. 
Newly added parameters are assigned $\tilde{\mF}=0$, so their effective learning rate is not attenuated ($\eta_{\mathrm{eff}}=\eta$), leaving them \emph{fully plastic after initialization}, providing fresh capacity for learning new tasks.
Orthogonal fan-in initialization encourages new neurons to begin from diverse input directions, ensuring \emph{minimal redundancy in new features}.
Finally, all hyperparameters are properties of the learning dynamics and \emph{not dependent on the task stream properties}, whether that is the number of tasks $T$, how similar they are to each other, or the task id itself.

\section{Theoretical Analysis} \label{sec:theory}

In fixed-capacity networks, continual learning is fundamentally constrained by the stability-plasticity dilemma: preserving prior knowledge (stability) inevitably limits the ability to acquire new tasks (plasticity), while unconstrained updates lead to catastrophic forgetting. 
Methods like \gls{ewc} work within this trade-off by protecting important parameters, but in doing so restrict the capacity available for learning future tasks. 
Consequently, even with such regularization, only a finite number of tasks can be learned without eventual plasticity decay.
Unlike \gls{ewc} or typical gradient descent, \gls{noracl} can circumvent the stability-plasticity trade-off through growth.

To theoretically illustrate this point, we consider an analytically tractable continual learning challenge originally used by \citet{kirkpatrick2017EWC}: random pattern association.
To enable growth (which requires a hidden layer), we extend this problem to a network with a single non-linear hidden layer initialized with width $M^{(0)}$.
Random pattern association presents a sequence of tasks indexed by $t$, where each task consists of associating a random binary input pattern, $u^{(t)}$, and its corresponding hidden representation, $\vx^{(t)}$, with a binary outcome, $y^{(t)}$. 
Given that the catastrophic forgetting problem under typical gradient descent has been well-explored, we limit our analysis to comparing how old and new task information is retained in \gls{ewc} and \gls{noracl}.

\begin{table}[t]
\centering
\caption{Comparison of NORACL and static baselines across multiple benchmarks. Boldface indicates the highest average accuracy within each benchmark and depth setting.}\vspace{0.2cm}
\label{tab:NORACL_static_detailed}
\small
\renewcommand{\arraystretch}{1.2}
\setlength{\tabcolsep}{4pt} %
\begin{tabular}{lcccccc}
\toprule
\multirow{2}{*}{\textbf{Model}} & \multicolumn{2}{c}{\textbf{Permuted MNIST}} & \multicolumn{2}{c}{\textbf{Rotated MNIST}} & \multicolumn{2}{c}{\textbf{Binary Split MNIST}} \\
\cmidrule(lr){2-3} \cmidrule(lr){4-5} \cmidrule(lr){6-7}
 & \textbf{Params} & \textbf{Acc (\%)} & \textbf{Params} & \textbf{Acc (\%)} & \textbf{Params} & \textbf{Acc (\%)} \\
\midrule
1L baseline small  & 12.7k & $44.0 \pm 0.9$ & 12.7k & $56.3 \pm 1.7$ & 12.7k & $71.6 \pm 1.9$ \\
1L baseline medium & 25.4k & $60.3 \pm 1.9$ & 25.4k  & $63.7 \pm 2.3$ & 25.4k & $\mathbf{72.8 \pm 0.9}$ \\
1L baseline large  & 50.8k & $76.0 \pm 0.8$ & 50.8k  & $\mathbf{73.6 \pm 1.2}$ & 50.8k  & $70.0 \pm 2.7$ \\
\textbf{1L NORACL} & 47.6k $\pm$ 1.6 & $\mathbf{79.9 \pm 0.5}$ & 42.2k $\pm$ 2.1 & $72.6 \pm 2.4$ & 23.8k $\pm$ 2.2 & $72.1 \pm 1.8$ \\
\midrule
2L baseline small  & 12.9k   & $41.9 \pm 1.3$ & 12.9k & $56.1 \pm 1.4$ & 12.9k & $72.1 \pm 1.8$ \\
2L baseline medium & 26.4k  & $56.5 \pm 1.1$ & 26.4k  & $63.3 \pm 1.7$ & 26.4k  & $71.2 \pm 2.6$ \\
2L baseline large  & 54.9k  & $73.3 \pm 1.8$ & 54.9k & $\mathbf{75.2 \pm 0.8}$ & 54.9k  & $72.5 \pm 1.8$ \\
\textbf{2L NORACL} & 49.2k $\pm$ 3.2 & $\mathbf{79.4 \pm 0.7}$ & 42.9k $\pm$ 2.7 & $74.9 \pm 1.7$ & 23.3k $\pm$ 3.1 & $\mathbf{73.9 \pm 2.5}$ \\
\bottomrule
\end{tabular}
\end{table}

\subsection{EWC maintains stability at the cost of future plasticity}

Under a \gls{mse} loss with \gls{ewc} regularization (full derivation in \cref{sec:appendix_binary_info_task}), the optimal output weights at task $t$ can be decomposed in terms of the knowledge about prior tasks stored in the old weights $\mW^{(t-1)}$, and the new information required to solve the current task expressed in the product of $\vx^{(t)}$, and $y^{(t)}$.

We then show that in a fixed capacity network, the proportion of new information being added decays with $t^{-1}$, i.e. inversely proportional to the number of tasks presented up to that point.
Effectively, \gls{ewc} trades off plasticity on new tasks for stability over old tasks, slowing down the rate of forgetting relative to unregularized gradient descent but not completely removing it with respect to newer tasks.
If we wanted our static network to be able to retain the entire sequence of random pattern-outcome pairs, we would need to allocate sufficient capacity \textit{a priori} (around $t \leq M^{(0)}$), introducing the need for a task oracle.

\subsection{NORACL circumvents stability-plasticity trade-off}

\gls{noracl} removes the need for a task oracle by dynamically increasing the network capacity as new tasks arrive, sidestepping the power-law plasticity decay of \gls{ewc} as well as the catastrophic forgetting of unregularized gradient descent on fixed-capacity networks.
Crucially, this enables it to handle new tasks (without any specific prior on the number of expected tasks) through continued growth while not overwriting existing task information.
We formalize these claims in the following theorem and its corollary.

\textbf{Theorem 1} (Stability of Parameters relevant to Previous Tasks)\\
Consider a network with a single hidden layer trained sequentially on random pattern association using \gls{noracl}.
Then, the parameters corresponding to previous learned tasks are strictly preserved at the same time as new parameters are added during growth.
That is, letting $\mW^{(t)}_{\text{old}} = \mW^{(t)}_{0:M^{(t-1)}}$
denote the subset of output parameters optimized for task $t$, but excluding the $k = M^{(t)} - M^{(t-1)}$ newly instantiated hidden dimensions,
$\mW^{(t)}_{\text{old}} = \mW^{(t-1)}$.

\textbf{Corollary 1} (Plastic Learning of new Tasks through new Parameters)\\
Under the same setting, the minimization of the loss for a new task $t$ is achieved exclusively through newly instantiated hidden units.

We provide the full proof in Appendix \cref{sec:appendix_binary_info_task}.
At a high level, the argument proceeds by demonstrating that both \gls{noracl} growth triggers are activated by new random input patterns, triggering an expansion of the hidden layer with each new task $t$.
This yields asymptotic linear growth of the hidden layer's size, $M^{(t)} \approx M^{(0)} + (1-\gamma) \cdot t$ where $\gamma$ is the \gls{ed} discount factor.
Note that this linear growth of capacity offsets the inverse scaling of plasticity in \gls{ewc} on fixed-capacity networks.
Given the linear increase in hidden layer size, we can decompose the gradient updates into contributions over old parameters (trained to solve all previous tasks $j < t$) and newly added parameters.
This decomposition then yields Theorem 1 and Corollary 1's central claims: previously learnt parameters are preserved, demonstrating the stability of old task information, while new parameters entirely absorb the information required to solve the new task, demonstrating continued plasticity.

In conclusion, this analysis on random pattern association highlights how \gls{noracl} resolves the stability-plasticity trade-off that plagues fixed-capacity networks without needing to resort to a task oracle. In \cref{{appendix:theory_transfer}}, we show that these insights translate qualitatively, from our simplified but analytically tractable, case to deep and nonlinear MLPs trained on permuted MNIST. By expanding its own capacity in proportion to the task demands and the number of tasks, and then routing new information into highly plastic, unconstrained parameters, it avoids both catastrophic forgetting and power-law plasticity decay.

\section{Empirical Evaluation} \label{sec:experiments}

We evaluate \gls{noracl} on three MNIST \citep{lecun2002gradient} variants spanning a spectrum of task geometry: \textbf{Permuted MNIST} (10 tasks; random pixel permutations, weakly-similar tasks \citep{goodfellow2015empirical}), \textbf{Rotated MNIST} (5 tasks; \citep{lopez2017gradient} fixed angular rotations of $20^\circ$ starting from $0^\circ$, moderately-similar), and \textbf{Binary Split MNIST} (5 tasks; binary classification over disjoint digit pairs similar to \cite{zenke2017continual}, highly similar). In addition, we evaluate \gls{noracl} on top of a frozen encoder on \textbf{Binary Split CIFAR-10} (5 tasks; binary classification over disjoint class pairs). All are framed as domain-incremental problems with a single shared output head and no task identity at inference (\Cref{sec:methods}).

For the MNIST benchmarks, we compare \gls{noracl} against static baselines with \gls{ewc} at hidden widths of 16, 32, and 64 neurons per layer, for both one- and two-layer \glspl{mlp}.
For Binary Split CIFAR-10, we follow \citet{de2026continual} to use a frozen, pre-trained ResNet18 as a fixed encoder and only train (or grow) the two-layer MLP head on top, comparing against static baselines with MLP widths swept from 128 to 1024.
\gls{noracl} uses identical growth hyperparameters ($\gamma$, $\varepsilon$, $p$, $C$, $s_{\mathrm{init}}$, $\alpha$) across all MNIST benchmarks and depths; Appendix \Cref{tab:hyperparams_appendix} lists the exact values for all the benchmarks. All runs use 5 seeds. Following \citet{mirzadeh2022wide}, we report average test accuracy across all tasks seen so far.

\subsection{Short-Horizon MNIST Benchmarks: Matching Oracle Capacity}
\label{sec:short_horizon}

\Cref{tab:NORACL_static_detailed} reports parameter counts and average accuracy after all the tasks are presented for every MNIST benchmark.

\textbf{\gls{noracl} performs similarly or better than the best static baseline while using fewer parameters.}
Across all three benchmarks and both depths, \gls{noracl} achieves accuracy on-par with (i.e.\ within error-bounds)  or greater than the largest static baseline while using 10-22\% fewer parameters. 
The gains are most pronounced on Permuted MNIST, where 2L \gls{noracl} outperforms the 2L-large baseline by $\sim$6\% at $\sim$10\% fewer parameters. 
On Binary Split MNIST, where task structure is strongly shared, all methods fall within $\sim$1.5\% of each other. Importantly, \gls{noracl} \emph{does not overgrow} on this task stream, matching the medium baseline at a comparable parameter budget.

\begin{figure*}[t]
    \centering
    \includegraphics[width=\textwidth]{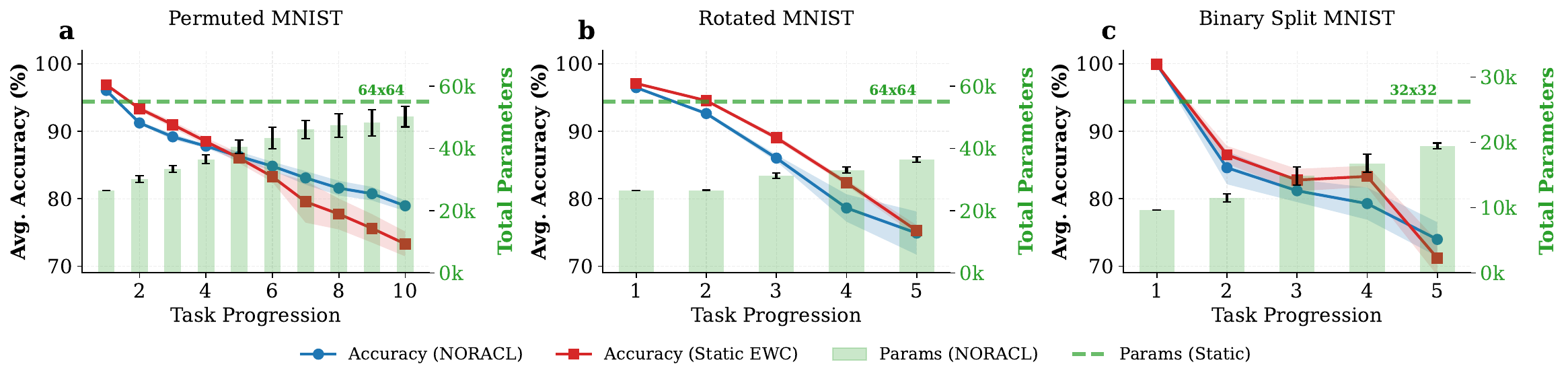}
    \caption{%
    \textbf{Accuracy \& parameter count vs.\ task progression on short-horizon benchmarks (2-layer models).} 
    Each panel shows average accuracy (left axis, curves) and total parameter count (right axis, bars) as a function of task progression for \gls{noracl} and the best-matching static \gls{ewc} baseline.
    The dashed green line indicates the static baseline's fixed parameter budget. 
    Error bars and shaded regions denote $\pm 1\sigma$ standard deviation over 
    5 seeds.}
    \label{fig:F2}
\end{figure*}

\textbf{\gls{noracl}'s parameter budget autonomously 
adapts to task geometry.} Despite using identical growth hyperparameters across all benchmarks, \gls{noracl}'s final parameter count varies by more than $2\times$ across task families: $\sim$49k on Permuted MNIST, $\sim$43k on Rotated MNIST, and only $\sim$23k on Binary Split MNIST. The growth trigger expands the network only when the current capacity appears insufficient, as indicated by the joint saturation of the spectral and Fisher-based signals. When tasks share structure, these signals rarely fire together and \gls{noracl} stays compact. When tasks are weakly related, they fire repeatedly and the network grows (see ablations in Appendix~\ref{sec:app_ablations}).

\Cref{fig:F2} visualizes average accuracy together with parameter usage over task progression for the 2-layer models. Across all benchmarks, \gls{noracl} either outperforms the matched static baseline at a comparable parameter budget, or achieves similar accuracy with fewer parameters throughout the task stream. This is consistent with the theoretical prediction (\Cref{sec:theory}) that growth escapes fixed-capacity ceilings on weakly-similar streams.

\subsection{Pretrained-feature CIFAR-10 benchmark}
\label{sec:cifar10}

\begin{wraptable}{r}{0.35\textwidth}
\centering
\small
\setlength{\tabcolsep}{4pt}
\caption{Binary Split CIFAR-10 (frozen ResNet18, 512-d features, MLP head).
Static EWC baselines span hidden widths $128$ to $1024$; NORACL grows from
$128\times128$ (averaged across $5$ seeds)}
\vspace{0.2cm}
\label{tab:cifar10}
\begin{tabular}{l r r@{\,$\pm$\,}l}
\toprule
Hidden width & Params & \multicolumn{2}{c}{Acc (\%)} \\
\midrule
\multicolumn{4}{l}{\textit{Static EWC}} \\
\quad $128\times128$           & 82.2k & 85.0 & 1.9 \\
\quad $256\times256$           & 197k  & 85.2 & 0.9 \\
\quad $512\times512$           & 525k  & 85.3 & 1.2 \\
\quad $1024\times1024$         & 1.5M  & 84.7 & 0.3 \\
\quad $172\times134^{\dagger}$ & 111k  & 85.4 & 0.8 \\
\midrule
\multicolumn{4}{l}{\textit{Adaptive}} \\
\quad NORACL & 105k$\pm$5.7k & \textbf{86.3} & \textbf{1.1} \\
\bottomrule
\end{tabular}
\\[0.3em]
{\footnotesize $^{\dagger}$Size-matched static control, provisioned from the start at NORACL's max grown size.}
\end{wraptable}

\Cref{tab:cifar10} reports parameter counts and average accuracy after all the tasks are presented for Binary Split CIFAR-10.

\textbf{Static baselines with \gls{ewc} do not improve with increased capacity.} Scaling the head from $82$k to $1.5$M parameters does not improve final accuracy ($85.0$--$85.4\%$), and the largest model is in fact marginally worse ($84.7\%$).

\textbf{\gls{noracl} performs similarly or better than the best static baseline while using fewer parameters.} The grown network matches and even slightly exceeds the best static baseline while growing (from a $128\times128$ initialization) to only $100\text{k}$ total parameters, i.e. ${\sim}5\times$ fewer than the $525$k network and ${\sim}14\times$ fewer than the $1.5$M network. It slightly improves on the size-matched static control ($172\times134$, $111$k) provisioned with the same final capacity from the start.

Together the results in \Cref{tab:cifar10} show that even when static models are given an order of magnitude more capacity on realistic features, they do not surpass a compact network that grows only as much as it needs, and provisioning that capacity up front is no better than allocating it adaptively.

\subsection{Emergent Architectures Reflect Task Geometry}
\label{sec:emergent_arch}

\begin{figure*}[t]
    \includegraphics[width=\textwidth]{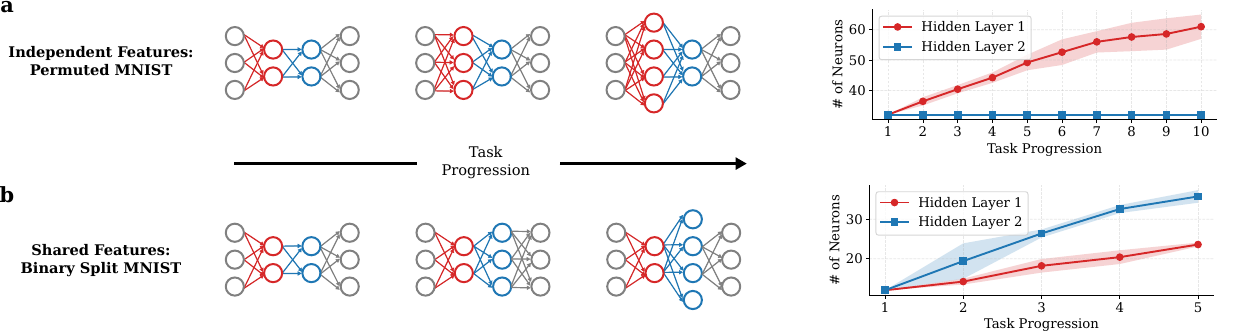}%

    \caption{%
    \textbf{Emergent architectures reflect task geometry.} 
    \textbf{Left:}~Schematic illustration of layer-wise 
    growth for independent-feature tasks (Permuted MNIST, \textbf{a)}) and shared-feature tasks (Binary Split MNIST, \textbf{b)}). 
    \textbf{Right:}~Number of neurons per hidden layer as a function of task progression for the 
    2-layer \gls{noracl} model.}
    \label{fig:F3}
\end{figure*}

\Cref{fig:F3} shows the number of neurons in each hidden layer as a function of task progression for a 2-layer \gls{noracl} model. The growth patterns differ strikingly across MNIST benchmarks, and each is interpretable in terms of the underlying task structure.

On Permuted MNIST, layer~1 grows steadily (from $\sim$32 to $\sim$60 neurons over 10 tasks) while layer~2 remains flat.
This is consistent with the nature of the task stream.
Random permutations destroy low-level spatial structure, such that each new task demands different first-layer filters, while the second layer's feature space remains largely reusable as long as the first layer continues to produce useful intermediate representations. 
\gls{noracl}'s growth trigger discovers this division of labor autonomously. 
Strikingly, on Binary Split MNIST, the pattern inverts: 
Layer~2 grows faster than layer~1 ($\sim$36 vs.\ $\sim$23). 
Here, all tasks share the same raw digit distribution, so low-level features are reusable and growth shifts to the decision boundary in Layer~2.
On Rotated MNIST (Appendix~\cref{fig:app_rmnist}), both layers grow, but layer~1 more than layer~2 ($\sim$44 vs.\ $\sim$37), reflecting that rotations alter pose while preserving some local statistics. 

Note that the same growth rule produces bottom-heavy growth on Permuted, balanced growth on Rotated, and top-heavy growth on Binary Split MNIST. We read this as evidence that \gls{noracl} is tracking a meaningful notion of \emph{where capacity is actually required}, rather than a surrogate that correlates with good accuracy. Consistent with this, the appendix shows that heuristic growth schemes such as fixed and scheduled expansion perform substantially worse, even though they also increase capacity (\Cref{sec:app_ablations}).

\subsection{Long-Horizon Stress Test: 50-Task Permuted MNIST}
\label{sec:long_horizon}

To expose the oracle architecture problem over a long task horizon, we simulate an extreme deployment scenario:
We predict a short stream of 5 tasks from Permuted MNIST and provision a small $32\times32$ network accordingly.
Unexpectedly however, the stream extends to 50 tasks.
We compare three conditions (5 seeds each): Static \gls{ewc} baseline $32\times32$ (the provisioned network), Static \gls{ewc} $75\times32$ (an architecture-matched control whose widths match \gls{noracl}'s max final size provisioned from the start), and \gls{noracl} initialized at $32\times32$ and allowed to grow.
\gls{ewc}. A fully unconstrained parameter contributes 1, a fully frozen parameter contributes 0, and partially constrained parameters contribute fractionally (for details see Appendix \Cref{sec:app_plasticity_diagnostics}).

\begin{wrapfigure}{r}{0.50\textwidth}
    \centering
    \includegraphics[width=0.50\textwidth]{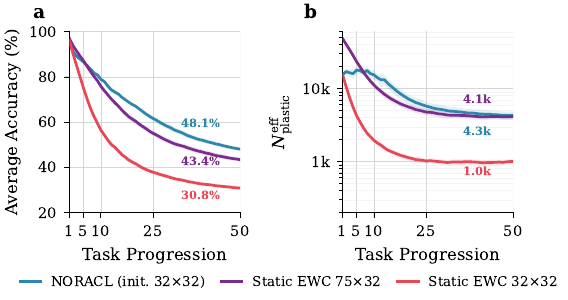}
    \caption{\textbf{50-task Permuted MNIST stress test.}
    \textbf{a)}~Average accuracy across all seen tasks.
    \textbf{b)}~Effective plastic parameter count  $N^{\mathrm{eff}}_{\mathrm{plastic}}$ (Eq.~\ref{eq:epp}, 
    log scale)}
    \label{fig:F4}
\end{wrapfigure}

\Cref{fig:F4}a reports the average accuracy over task progression across 50 tasks.
For all three models, accuracy drops with increasing task number, but in qualitatively different ways.
The static $32\times32$ network deteriorates fastest, consistent with rapid exhaustion of its under-provisioned fixed capacity.
The architecture-matched static $75\times32$ baseline initially benefits from starting with the full pre-provisioned capacity, but even so is overtaken by \gls{noracl} after less than 10 tasks.
Overall, \gls{noracl} finishes strongest, remaining the most stable in terms of average accuracy across all 50 tasks despite reaching a comparable final size to the architecture-matched baseline.

\Cref{fig:F4}b shows that the effective number of plastic parameters ($N^{\mathrm{eff}}_{\mathrm{plastic}}$) decays slowest for \gls{noracl} when compared to the static networks.
The static $32\times32$ network deteriorates fastest, consistent with rapid exhaustion of its fixed capacity.
The architecture-matched static $75\times32$ baseline starts with far more total parameters than either \gls{noracl} or the static $32\times32$ network, explaining its initial advantage in effective plastic parameters.
However, as with the accuracy, \gls{noracl} overtakes it after only 10 tasks (even before reaching its final size), maintaining a small plasticity advantage as more tasks accumulate.

These results directly illustrate how \gls{noracl} improves upon the stability-plasticity dilemma through growth. 
In this sense, the long-horizon stress test is not only a comparison of accuracies, but a mechanistic demonstration of how adaptive neurogenesis can move the stability-plasticity Pareto frontier by allocating new resources only when the current architecture becomes genuinely insufficient.
The fact that \gls{noracl} even outperforms a static architecture pre-provisioned with the same final capacity demonstrates that \textit{how} and \textit{when} growth occurs is at least as important as \textit{that} it occurs.

One limitation is that \gls{noracl}'s final accuracy of $\sim$48\% on 50-task Permuted MNIST is not a \emph{high} absolute number, the comparison is only meaningful \emph{relative} to the static $32\times32$ baseline that starts from the same capacity and the capacity matched $75\times32$ baseline. We are aware that a large static oracle baseline would likely outperform both given the knowledge about the task count. 

\subsection{Ablation Studies}\label{sec:ablation_study}

In Appendix~\ref{sec:app_ablations}, we ablate both components of \gls{noracl}'s growth mechanism: the joint ED~$\wedge$~$F_{\mathrm{sat}}$ trigger (\Cref{sec:app_ablations_growth_trig}) and the QR-based orthogonal initialization (\Cref{sec:app_ablations_init}).
We find that (i)~\gls{ed} alone leads to overgrown networks with an average $150\text{k}$ more parameters and $1.2\%$ lower accuracy across tasks, while (ii)~$F_{\mathrm{sat}}$ alone misallocates capacity across layers and results in an average $4.2\%$ lower accuracy across tasks.
Note that using \gls{ed} alone is equivalent to deploying NORTH* \citep{maile2022and} per task while simultaneously using \gls{ewc} regularization, serving as a natural expansion-based baseline. In addition, \Cref{sec:app_ablations_growth_trig} shows that our joint ED~$\wedge$~$F_{\mathrm{sat}}$ trigger consistently outperforms other growth heuristics (by at least $1.9\%$) including fixed growth per task, scheduled growth, and loss-based growth as used by \citet{yoon2017lifelong}.
In terms of initializations, our QR initialization consistently produces the most compact architectures at the highest accuracy and outperforms other common initializations (by at least $1.3\%$).

\section{Discussion \& Conclusion} \label{sec:discussion}
In this work, we argue that the stability-plasticity trade-off in \gls{cl} has an architectural root. 
Fixed-capacity methods implicitly rely on an oracle architecture sized for an unknown future task stream, as the required network capacity depends on both task count and task geometry.
\gls{noracl} addresses this oracle architecture problem through smart neurogenesis: it starts from a compact network and grows only when and where needed, guided by internal signals of representational- and plasticity-saturation.
Our theoretical results clarify how this is possible: Theorems~1 and Corollary 2 show, in an analytically tractable setting, that \gls{noracl} preserves parameters relevant to previous tasks while routing new learning into newly added parameters, thereby avoiding the power-law plasticity decay of fixed-capacity regularized networks. 
This mechanism is also reflected empirically in our 50-task stress test, where \gls{noracl} maintains a substantially larger pool of effective plastic parameters than \gls{ewc} networks and degrades more gracefully as tasks accumulate (\cref{fig:F4}). 
In this way, \gls{noracl} moves the stability-plasticity Pareto frontier by allocating additional resources only when the current architecture is genuinely insufficient. 
This, and the growth's independence of task information, also distinguishes it from earlier expansion-based methods such as Progressive Neural Networks~\citep{rusu2016progressive} and Dynamically Expandable Networks~\citep{yoon2017lifelong}, which rely on task identity at inference time and use heuristic growth criteria, making them less suitable for the domain-incremental setting considered here.

A particularly interesting outcome is that the architectures learned by \gls{noracl} are both minimal on demand and grow in an interpretable fashion. Across benchmarks, \gls{noracl} reaches final accuracies on par with oracle-provisioned static baselines while using 10-20\% fewer parameters, showing that it can discover architectures whose size is close to what is actually needed rather than what must be pessimistically provisioned in advance (\cref{fig:F2}).
Moreover, the layer-wise growth patterns reflect task geometry in an intuitive way: when tasks are weakly related or largely orthogonal, growth is concentrated in earlier feature-extraction layers, consistent with the need to allocate new low-level features that cannot be shared.
When tasks share more structure, growth shifts toward later feature-combination layers, suggesting that early features can be reused and only downstream recombination must expand (\cref{fig:F3}). This makes \gls{noracl} interesting not only as a continual learning method, but also as a potential form of online architecture search.
In future work, such adaptive growth could be especially valuable in resource-constrained settings, including adaptive hardware systems, where one may need to decide on the fly how much memory, compute, or circuitry to allocate.

In summary, we identify the oracle architecture problem as a central cause of the stability-plasticity dilemma and show that smart, on-demand neurogenesis offers a principled way to address it. More broadly, our results suggest that continual learning systems should adapt not only their parameters, but also their capacity. In this sense, \gls{noracl} provides a concrete step toward pushing the stability-plasticity Pareto frontier of continual learning.

\paragraph{Limitations and future work:} \label{subsec:limits}
\gls{noracl} is currently formulated for fully-connected \gls{mlp} architectures. Extending it to convolutional or attention-based models, where growth means adding filters or attention heads rather than neurons, i3s conceptually straightforward (the \gls{ed} and Fisher signals generalize naturally to feature maps) but requires extensive engineering and validation on benchmarks beyond MNIST and CIFAR-10. We view the present work as establishing the \emph{principle} that joint spectral-and-plasticity gating resolves the oracle architecture problem and that scaling to richer architectures, such as convolutional neural networks, is the natural next step. Additionally, our evaluation deliberately uses small- and medium-sized networks, because the oracle architecture problem and the associated loss of plasticity are most acute in the capacity-constrained regime. In larger networks, plasticity exhaustion still occurs but over longer task horizons~\citep{mirzadeh2022wide}, making it harder to expose at MNIST scale.

\section*{Acknowledgments}

This work was supported by the Swiss National Science Foundation (Starting Grant Project UNITE, grant number TMSGI2-211461), the Horizon Europe program (EIC Pathfinder METASPIN, grant number 101098651) and VolkswagenStiftung (project CLAM, grant number 9C854).
The authors appreciate the support provided by the project funding and institutional resources that made this research possible.
We thank all members of the Emerging Intelligent Substrates Lab (EIS-Lab) for their valuable feedback and support at multiple stages of this project. We are particularly grateful to Filippo Moro for suggesting the frozen convolutional stem with a grown MLP head used in our CIFAR-10 experiments. We also thank Damien Querlioz, Djohan Bonnet, Kellian Cottart, and Emre Neftci for insightful discussions and feedback. Finally, we thank Mathis Reymond for his contributions to the preliminary version of this project.
\bibliography{collas2026_conference}
\bibliographystyle{collas2026_conference}

\newpage
\appendix
\section{Appendix}
\subsection{Implementation Details of NORACL}\label{sec:app_implementation_details}

\subsubsection{Additional details on how NORACL adds neurons.}

When \gls{noracl} adds $k_l$ neurons to layer $l$, we draw a random matrix $\mA \in \mathbb{R}^{M_{l-1} \times M_{l-1}}$, compute its QR decomposition $\mA = \mQ\mR$~\citep{saxe2013exact}, and take the first $k_l$ columns of $\mQ$ as the new fan-in weights. In the baseline implementation, these columns are scaled by a fixed factor $s_{\mathrm{init}}$: \begin{equation} \label{eq:fanin_init_combined} \mW_l^{(\mathrm{new})} = s_{\mathrm{init}}\, \mQ_{:,\,1:k_l}. \end{equation} This yields orthogonal incoming weight vectors with controlled magnitude. In our experiments, $s_{\mathrm{init}}$ is treated as a hyperparameter (set to $0.2$). The corresponding fan-out weights into layer $l+1$ are initialized to zero, 
\begin{equation} 
\label{eq:fanout_init_combined} 
\mW_{l+1}^{(\mathrm{new})} = \mathbf{0}, 
\end{equation} 
\begin{equation} 
\label{eq:fanout_init_combined} 
\mW_{l+1}^{(\mathrm{new})} = \mathbf{0} \in \mathbb{R}^{k_l \times M_{l+1}}. 
\end{equation} 
so that the newly added neurons do not affect downstream computation at the moment of insertion. This makes growth function-preserving, i.e.\ $f(\vx;\theta') = f(\vx;\theta)$ for all inputs $x$ immediately after expansion.
The accumulated Fisher diagonal and the \gls{ewc} anchor parameters are padded with zeros for the new dimensions to ensure that new parameters remain fully unconstrained.

\begin{equation} 
\tilde{\mF}_l \leftarrow [\tilde{\mF}_l \mid \mathbf{0}], \quad \tilde{\mF}_{l+1} \leftarrow \begin{bmatrix} \tilde{\mF}_{l+1} \\ \mathbf{0} \end{bmatrix}, \quad \theta^*_l \leftarrow [\theta^*_l \mid \mathbf{0}], \quad \theta^*_{l+1} \leftarrow \begin{bmatrix} \theta^*_{l+1} \\ \mathbf{0} \end{bmatrix}. 
\end{equation} 
Setting the Fisher information to zero for new parameters means they receive no \gls{ewc} penalty (see \cref{eq:grohess_oewc_loss}), making them fully plastic. 

\subsubsection{Pseudocode}

\begin{algorithm}[h] 
\caption{\gls{noracl}: Neurogenesis for Oracle-free Resource-Adaptive Continual Learning} 
\label{alg:grohess} 
\begin{algorithmic}[1] 
\REQUIRE Initial network $\theta_0$ with hidden widths $M_1^{(0)}, \ldots, M_L^{(0)}$; task stream $\{\mathcal{D}_t\}_{t=1,\ldots,T}$ 
\REQUIRE Hyperparameters: $\lambda, \alpha, \gamma, \varepsilon, p, C, s_{\mathrm{init}}$ \hfill \textit{(no dependence on $T$)} 
\STATE Initialize $\tilde{F} \leftarrow \mathbf{0}$, $\theta^* \leftarrow \theta_0$, cooldown $\leftarrow 0$, and $\varphi_l^{(0)} \leftarrow \emptyset$ 
\FOR{each arriving task $\mathcal{D}_t$} 
    \FOR{epoch $= 1$ to $E$}           
        \FOR{each mini-batch $(x, y) \in \mathcal{D}_t$} 
            \STATE $\mathcal{L} \leftarrow \mathcal{L}_{\mathrm{CE}}(\theta; x, y) + \frac{\lambda}{2} \sum_i \tilde{F}_i (\theta_i - \theta_i^*)^2$ \hfill \textit{(Eq.~\ref{eq:grohess_oewc_loss})} 
            \STATE $\theta \leftarrow \theta - \eta \, \nabla_\theta \mathcal{L}$ \hfill \textit{(SGD with gradient clipping)}
        \ENDFOR 
        \IF{cooldown $> 0$} 
            \STATE cooldown $\leftarrow$ cooldown $- 1$ 
        \ELSE 
            \STATE Compute $\varphi_l$ for each hidden layer $l$ \hfill \textit{(Eq.~\ref{eq:ed})} 
            \STATE Estimate current-task Fisher $F_l^{(\mathrm{curr})}$ for each layer using the first few mini-batches of $\mathcal{D}_t$ 
            \STATE growth\_happened $\leftarrow \textbf{false}$ 
            \FOR{each hidden layer $l$} 
                \IF{$\varphi_l > \gamma \cdot \varphi_l^{(0)}$ \textbf{and} $\mathrm{Percentile}(F_l^{(\mathrm{curr})}, p) > \tau_l$} 
                    \STATE $k_l \leftarrow \max(1,\; \lfloor M_l (\varphi_l - \gamma \varphi_l^{(0)}) \rfloor)$ 
                    \STATE $\theta, \tilde{F}, \theta^*, \texttt{growth\_happened} \leftarrow \textsc{Grow}(\theta, \tilde{F}, \theta^*, l, k_l, s_{\mathrm{init}})$ \hfill \textit{(\Cref{sec:method_growth})} 
                \ENDIF 
            \ENDFOR 
            \IF{growth\_happened} 
                \STATE cooldown $\leftarrow C$ 
                \STATE Reinitialize optimizer state on the expanded parameter tree 
            \ENDIF 
        \ENDIF 
    \ENDFOR 
    \STATE \textit{// Post-task consolidation} 
    \STATE $F_t \leftarrow \mathrm{compute\_fisher}(\theta, \mathcal{D}_t^{(\leq 5\ \text{mini-batches})})$ 
    \STATE $\tilde{F} \leftarrow \alpha \tilde{F} + (1-\alpha) F_t$ \hfill \textit{(Eq.~\ref{eq:fisher_blend})} 
    \STATE $\theta^* \leftarrow \theta$ 
    \STATE $\tau_l \leftarrow \alpha \, \tau_l + (1-\alpha) \, \mathrm{mean}(\tilde{F}_l)$ for each layer $l$ 
    \STATE $\varphi_l^{(0)} \leftarrow \mathrm{orth\_score}_l / M_l$ for each hidden layer $l$ \hfill \textit{(set next task's reference ED)} 
\ENDFOR 
\end{algorithmic} 
\end{algorithm}

\FloatBarrier

\subsubsection{Hyperparameter Settings}
\label{app:hyperparams}

\begin{table}[t]
\centering
\begin{threeparttable}
\caption{Hyperparameter settings used in all experiments. NORACL uses the same growth-related hyperparameters across all three MNIST benchmarks and both depths. For Binary Split CIFAR-10, the growth-trigger thresholds ($\gamma$, $p$) are adjusted to reflect a property of the feature distribution (see \cref{sec:cifar10}).}
\label{tab:hyperparams_appendix}
\renewcommand{\arraystretch}{1.15}
\setlength{\tabcolsep}{6pt}
\begin{tabular}{llcc}
\toprule
\textbf{Component} & \textbf{Hyperparameter} & \textbf{MNIST-based} & \textbf{Split CIFAR-10} \\
\midrule
\multirow{6}{*}{\textbf{NORACL growth}}
& ED discount & $\gamma = 0.9$ & $\gamma = 0.8$ \\
& SVD threshold & $\varepsilon = 0.05$ & $\varepsilon = 0.05$ \\
& Fisher percentile & $p = 25$ & $p = 50$ \\
& Cooldown epochs & $C = 3$ & $C = 3$ \\
& Init scale & $s_{\mathrm{init}} = 0.2$ & $s_{\mathrm{init}} = 0.2$ \\
& Fisher EMA decay & $\alpha = 0.9$ & $\alpha = 0.9$ \\
\midrule
\multirow{3}{*}{\textbf{EWC}}
& Permuted MNIST & $\lambda = 500$ & \multirow{3}{*}{$\lambda = 5000$} \\
& Rotated MNIST & $\lambda = 2000$ & \\
& Binary Split MNIST & $\lambda = 5000$ & \\
\midrule
\multirow{4}{*}{\textbf{Optimization}}
& Optimizer & SGD & SGD \\
& Learning rate (task 1) & $\eta_1 = 1\times 10^{-1}$ & $\eta_1 = 1\times 10^{-1}$ \\
& Learning rate (later tasks) & $\eta = 5\times 10^{-3}$ & $\eta = 5\times 10^{-3}$ \\
& Gradient clipping & $5.0$ & $5.0$ \\
\midrule
\multirow{4}{*}{\textbf{Training}}
& Seeds & 5 (0-4) & 5 (0-4) \\
& Batch size & $256$ & $256$ \\
& Epochs (task 1) & $10$ & $10$ \\
& Epochs (later tasks) & $30$ & $80$ \\
\bottomrule
\end{tabular}
\end{threeparttable}
\end{table}

\FloatBarrier

\subsection{Proofs and Extensions to Theoretical Analysis}

\subsubsection{Task Information Retention for Binary Association Task}\label{sec:appendix_binary_info_task}

To analytically understand the representational capacity of networks trained with the \gls{noracl} algorithm, we analyze the dynamics of task information retention over time on the simple (but analytically tractable) task of associating an arbitrary number of random binary strings to random binary outputs as done by \citet{kirkpatrick2017EWC}.
From a continual learning perspective, we consider the problem of memorizing each new $(\vx^{(t)},y^{(t)})$ pair as a new task indexed by $t$.

To solve this task, we train networks with a single hidden layer of size $M$. The network must associate a $d$-dimensional input string $\vu^{(t)}$ to a binary output $y^{(t)}$ at every timestep $t$. To evaluate the effects of growing the network via \gls{noracl} while avoiding the representational bottlenecks of strictly linear networks, we ground our analysis in a Random Feature framework. 
We define the $M$-dimensional hidden activations as a deterministic, non-linear projection of the input: $\vx^{(t)} = \sigma(\mQ\vu^{(t)})$.
Here, $\mQ \in \mathbb{R}^{M \times d}$ is a random projection matrix with entries drawn from $\mQ_{j,k} \sim \mathcal{N}(0,1)$, and $\mW \in \mathbb{R}^{1 \times M}$ represents the output weights.
Because of the non-linearity $\sigma$, expanding the hidden layer size $M$ increases the representational capacity of the network. 

Assuming inputs are normalized such that $||\vu^{(t)}||_2=1$, pre-activations are distributed as standard normal variables.
We restrict our choice of the non-linearity $\sigma$ such that it has zero mean and unit variance over the standard Gaussian measure: $\mathbb{E}_{z \sim \mathcal{N}(0,1)}[\sigma(z)] = 0$ and $\mathbb{E}_{z \sim \mathcal{N}(0,1)}[\sigma(z)^2] = 1$. 
Accordingly, we can also model the post-activations $\vx^{(t)}$ as independent, zero-mean features. We learn $\mW$ using least-squares between the high-dimensional features $\vx^{(t)}$ and targets $y^{(t)}$ in addition to an \gls{ewc} regularization penalty, yielding the following loss for timestep $t$ (following the initial derivation by \citet{kirkpatrick2017EWC}):

\begin{equation}\label{eq:mse_ewc_loss}
L^{(t)}=\frac{1}{2}(||\mW^{(t)}\vx^{(t)}-y^{(t)}||_2^2+|\mW^{(t)}-\mW^{(t-1)}|_\mathcal{M})
\end{equation}

Here, $\mW^{(t)}$ refers to the output weights at timestep $t$, while $|\mW^{(t)}-\mW^{(t-1)}|_\mathcal{M}$ refers to a distance penalty (under metric $\mathcal{M}$) between the weights learned at the time of the previous task $t-1$ and the current weights at time $t$. For \gls{ewc}, the metric $\mathcal{M}$ is defined by the cumulative sum of the diagonal of the Fisher information matrix up to timestep $t-1$.

\subsubsection*{Case 1: Elastic Weight Consolidation on a Static-Sized Network}

We first consider the case of training a neural network with a static size $M^{(0)}$ using \gls{ewc} without growth triggers. 
The following derivation simply reproduces the steps outlined in the supplementary information of \citet{kirkpatrick2017EWC}.
In this static case, the diagonal of the Fisher information matrix for all tasks up to $t$, denoted denote $\mF^{(t)}$, is simply the sum of Fisher diagonals $\mF_j$ for each individual task $j \in 0,...,t$ (what the Fisher would be if the network was only trained to solve task $j$).

To appropriately control the variance of the forward pass in this derivation, we adopt the standard scaling assumption from \citet{kirkpatrick2017EWC} and scale the pre-activations by $\frac{1}{\sqrt{M^{(0)}}}$, such that $\vx^{(t)} = \sigma(\frac{1}{\sqrt{M^{(0)}}}\mQ\vu^{(t)})$. 
Under this scaling, the expected Fisher matrix for a single task is $\mF_j = \mathbb{E}[\vx^{(j)} \vx^{(j)T}] = \frac{1}{M^{(0)}}\mI$. Consequently, the accumulated metric applied at timestep $t$ is $\mF^{(t)} = \frac{t}{M^{(0)}}\mI$. Letting $\mF^{(t)}=\lambda^{(t)} = \frac{t}{M^{(0)}}$, we can simplify \cref{eq:mse_ewc_loss} to the following:

\begin{equation}\label{eq:static_mse_ewc_loss}
L^{(t)}=\frac{1}{2}(||\mW^{(t)}\vx^{(t)}-y^{(t)}||_2^2 + \lambda^{(t)} ||\mW^{(t)}-\mW^{(t-1)}||_2^2)
\end{equation}

Minimizing the loss at timestep $t$ with respect to $\mW^{(t)}$ and setting the derivative to zero yields the following weight dynamics:

\begin{equation}\label{eq:weight_dynamics}
\mW^{(t)}=\mW^{(t-1)} - \frac{1}{\lambda^{(t)}}(\mW^{(t)}\vx^{(t)}-y^{(t)})\vx^{(t)T}
\end{equation}

We now solve for $\mW^{(t)}$ by applying the Sherman-Morrison formula for the resulting matrix inverse (recovering the dynamics shown in \citet{kirkpatrick2017EWC}):

\begin{align}\label{eq:weight_dynamics_solved}
\mW^{(t)}&=\mW^{(t-1)} \left( \mI - \vx^{(t)}\frac{y^{(t)T}\vx^{(t)T}}{\lambda^{(t)} + 1} \right) + \frac{y^{(t)T}\vx^{(t)T}}{\lambda^{(t)} + 1} \\ &=\mW^{(t-1)} \left( \mI - \vx^{(t)}\frac{y^{(t)T}\vx^{(t)T}}{\frac{i}{M^{(0)}} + 1} \right) + \frac{y^{(t)T}\vx^{(t)T}}{\frac{i}{M^{(0)}} + 1}
\end{align}

We can interpret the first term of this equation as the proportion of knowledge of past tasks stored in $\mW^{(t-1)}$ that is retained, while the second term describes the new information about task $t$ being added to the weight matrix over time. 
We see that for static networks with \gls{ewc} regularization, the proportion of new task information that can be stored inevitably decays as a power law with respect to the number of encountered tasks $t$ as long as the network capacity $M^{(0)}$ is fixed.

\subsubsection*{Case 2: Dynamic Growth via the NORACL Algorithm}

We now turn to the second case, a network that is dynamically grown using the \gls{noracl} algorithm. We start by showing how a new task $t$ in this simplified framework guarantees that \gls{noracl} will expand the size of the hidden vector $\vx^{(t)}$. 
In this dynamically growing case, the size of $\vx^{(t)}$, denoted $M^{(t)}$, is a function of how many tasks have been seen.

Specifically, we will outline a line of reasoning that serves as proof to Theorem 1 and Corollary 1 in \cref{sec:theory}, restated here for convenience.

\textbf{Theorem 1} (Stability of Parameters relevant to Previous Tasks)\\
Consider a network with a single hidden layer trained sequentially on random pattern association using \gls{noracl}.
Then, the parameters corresponding to previous learned tasks are strictly preserved at the same time as new parameters are added during growth.
That is, letting $\mW^{(t)}_{\text{old}} = \mW^{(t)}_{0:M^{(t-1)}}$
denote the subset of output parameters optimized for task $t$, but excluding the $k = M^{(t)} - M^{(t-1)}$ newly instantiated hidden dimensions,
$\mW^{(t)}_{\text{old}} = \mW^{(t-1)}$.

\textbf{Corollary 1} (Plastic Learning of new Tasks through new Parameters)\\
Under the same setting, the minimization of the loss for a new task $t$ is achieved exclusively through newly instantiated hidden units.

Our proof proceeds by showing that the conditions for growth are met asymptotically by every random pattern presented to the network, then deriving the optimal weights for associating a new random feature, $\vx^{(t)}$, to its label $y^{(t)}$, and finally showing how these optimal weights change (in the presence of \gls{ewc} regularization) as new tasks are presented and growth occurs.

There are two conditions for \gls{noracl} to add new neurons: first, the Fisher saturation indicator must be true, and second, the effective dimensionality must exceed the discounted previous effective dimensionality.

Because $\mF^{(t)} = \mF^{(t-1)} + \mF_t$ and the MSE is a convex loss function ($\mF_t \geq \mathbf{0}$), it is trivially true that $\mF^{(t)} \geq \mF^{(t-1)}$. 
Given uniformly distributed random samples of $\vu^{(t)}$ without repetition (such that the network cannot already be at a global minimum at timestep $t$ and such that in expectation, $\mF^{(i)}=\mF^{(j)}$ for any $i,j \leq T$), $\mF^{(t)}$ will be strictly larger than $\mF_{(t-1)}$, and given an appropriate Fisher percentile setting, $\mF^{(t)} > \tau^{(t)}$. 
Therefore, given the correct setting for the , the Fisher saturation indicator will be true for any new task $t$. 

Given any task $t$ defined by input $\vu^{(t)}$ and output $y^{(t)}$, the effective dimensionality of $\vx^{(t)}$ will be $\phi^{ED}(t) = \frac{1}{M^{(t-1)}}$ (given a small enough $\epsilon$ threshold) since we perform singular value decomposition over the hidden values generated by a single input vector $\vu^{(i)}$ before growing the network. Since the discount factor $\gamma < 1$, it holds that $\phi^{ED}(t) > \gamma\phi^{ED}(t-1)$ for any new task $t$. 
Consequently, \gls{noracl} will add on average $M^{(t)}-M^{(t-1)} = M^{(t-1)} \cdot (\phi^{ED}(t) - \gamma\phi^{ED}(t-1)) \approx (1-\gamma)$ new neurons for every newly introduced task, assuming that for large $M$, $\frac{1}{M^{(t)}} \approx \frac{1}{M^{(t-1)}}$. We can thus approximate the size of the hidden layer $M^{(t)}$ as a linear function of the task number $i$ and the initialization size $M^{(0)}$:

\begin{equation}\label{eq:size_scaling}
M^{(t)} \approx M^{(0)} + (1-\gamma) \cdot t
\end{equation}

To understand \cref{eq:mse_ewc_loss} for networks trained using \gls{noracl}, the Fisher diagonal $\mF_{0:t}$ deserves careful attention. Unlike the static network case, the penalty $|\mW^{(t)}-\mW^{(t-1)}|_\mathcal{M}$ is now ill-defined because the dimensions of $\mW^{(t)}$ and $\mW^{(t-1)}$ no longer match. We address this by padding $\mW^{(t-1)}$ with future units to match the new dimension of $\mW^{(t)}$. Specifically, we concatenate $k = M^{(i)}-M^{(i-1)}$ zero-entries: $\tilde{\mW}^{(t-1)} = \begin{bmatrix}\mW^{(t-1)} & \mathbf{0}_{k}\end{bmatrix}$. 

We then define the growth-augmented Fisher diagonal to mask out the distance penalty for these newly added dimensions. We can express the growth-augmented Fisher diagonal $\bar{\mF}^{(t)}$ in terms of the Fisher $\mF_j$ for each individual task $j\leq t$ as follows:

\begin{equation}
\bar{\mF}^{(t)} = \sum_{j=0}^t \begin{bmatrix}\mF_{j} & \mathbf{0}_{M^{(i)}-M^{(j)}}\end{bmatrix} = \sum_{j=0}^i \begin{bmatrix}\mI_{M^{(j)}} & \mathbf{0}_{M^{(i)}-M^{(j)}}\end{bmatrix}
\end{equation}

Let $\mLambda^{(t)} = bar{\mF}^{(t)}$ denote the accumulated penalty metric used at timestep $t$. $\mLambda^{(t)}$ is a diagonal matrix where the entry for the $m$-th parameter, $\lambda^{(t)}_{m}$, represents its accumulated Fisher information. $\mLambda^{(t)}$ partitions all parameters into two sets:
1. For "old" parameters introduced in some past task $j$, $\lambda^{(t)}_{m} = t-j$.
2. For the $k$ "new" parameters added by \gls{noracl} specifically for task $t$, $\lambda^{(t)}_{m} = 0$.

To make $\mLambda^{(t)}$ invertible for our weight dynamics derivation, we assume a negligibly small isotropic prior $\epsilon \mI$ on all weights (where $\epsilon \to 0$), such that new weights have $\lambda^{(t)}_{m} = \epsilon$. Our modified loss function for \gls{noracl} at task $t$ becomes:

\begin{equation}\label{eq:mse_grohess_loss}
L_t = \frac{1}{2} \left( ||\mW^{(t)}\vx^{(t)}-y^{(t)}||_2^2 + (\mW^{(t)} - \tilde{\mW}^{(t-1)})^T \mLambda_t (\mW^{(t)} - \tilde{\mW}^{(t-1)}) \right)
\end{equation}

Minimizing \cref{eq:mse_grohess_loss} with respect to $\mW^{(t)}$ and setting the derivative to zero yields:

\begin{equation}\label{eq:grohess_weight_gradient}
0 = (\mW^{(t)}\vx^{(t)} - y^{(t)})x^{(t),T} + (\mW^{(t)} - \tilde{\mW}^{(t-1)})\mLambda^{(t)}
\end{equation}

Isolating $\mW^{(t)}$ and applying the Sherman-Morrison formula for the matrix inverse results in the exact weight dynamics for the growing network:

\begin{align}\label{eq:grohess_weight_dynamics_solved}
\mW^{(t)} &= \tilde{\mW}^{(t-1)} \left( \mI - \vx^{(t)}\frac{ \vx^{(t),T} \mLambda^{(t),-1}}{1 + \vx^{(t),T} \mLambda^{(t),-1} \vx^{(t)}} \right) + y_t\frac{\vx^{(t),T} \mLambda^{(t),-1}}{1 + \vx^{(t),T} \mLambda^{(t),-1} \vx^{(t)}} \\
&= \tilde{\mW}^{(t-1)} \left( \mI - \vx^{(t)}p^{(t),T} \right) + y_i\vp^{(t),T}
\end{align}

Note that we define $\vp^{(t)}$ such that it is a column vector with one entry for each of the $M^{(i)}$ parameters. Analyzing the denominator of $\vp^{(t)}$ shared across $M^{(i)}$ entries, we see:

\begin{equation}
1 + \vx^{(t),T} \mLambda^{(t),-1} \vx^{(t)} = 1 + \sum_{m=1}^{M^{(i)}} \frac{(x_m^{(t)})^2}{\lambda^{(t)}_{m}}
\end{equation}

Because the $k$ newly added dimensions have $\lambda^{(t)}_{m} = \epsilon \to 0$, their corresponding terms in the sum approach infinity. Consequently, the entire denominator approaches infinity. We can now observe the distinct, decoupled effect this has on the entries of $\vp^{(t)}$ (indexed by $m<M^{(i)}$) associated with the "old" versus "new" hidden units. 

For "old" hidden units (where $\lambda^{(t)}_{m}$ is a finite, non-zero value accumulating over past tasks and $m\leq M^{(t-1)}$), each $p^{(t)}_m$ evaluates to:

\begin{equation}
p^{(t)}_m = \lim_{\epsilon \to 0} \frac{x_m^{(t)} / \lambda^{(t)}_{m}}{1 + \sum_{l=1}^{M^{(t-1)}} \frac{(x_l^{(t)})^2}{\lambda^{(t)}_{l}} + \sum_{l=M^{(t-1)}}^{M^{(t)}} \frac{(x_l^{(t)})^2}{\epsilon}} = 0
\end{equation}

Letting $\mW^{(t)}_{\text{old}}=\mW^{(t)}_{:,0:M^{(i-1)}}$ and noting that $\tilde{\mW}^{(t-1)}_{\text{old}}=W^{(t-1)}$, we substitute $\vp^{(t)}_{0:M^{(i-1)}}=\begin{bmatrix}0&0&...&0\end{bmatrix}^T$ into \cref{eq:grohess_weight_dynamics_solved} and see that the contribution of the new task $i$ to the old weights vanishes. 

\begin{equation}\label{eq:grohess_weight_dynamics_solved}
\mW^{(t)}_{\text{old}}
= \tilde{\mW}^{(t)}_{\text{old}} \left( \mI - \vx^{(t)}_{:,0:M^{(i-1)}}\begin{bmatrix}0&0&...&0\end{bmatrix}^T \right) + y_i\begin{bmatrix}0&0&...&0\end{bmatrix}^T=\mW^{(t-1)}
\end{equation}

Thus, the parameters learned for previous tasks are preserved perfectly even as new tasks are added: $\mW^{(t)}_{\text{old}} = \mW^{(t-1)}$.
This concludes the proof for Theorem 1.

To prove Corollary 1, we continue by analyzing what happens to the recently added neurons.
For these "new" hidden units (where $\lambda^{(t)}_{m} = \epsilon$ and $m>M^{(t-1)}$), we see the following as $\epsilon \to 0$:

\begin{equation}
p^{(t)}_m  = \frac{x_m^{(t)}}{\sum_{l=M^{(t-1)}}^{M^{(t)}} (x_l^{(t)})^2}
\end{equation}

The fraction evaluates to a non-zero constant. 
Letting $\mW^{(t)}_{\text{new}}=\mW^{(t)}_{:,M^{(i-1)}:M^{(i)}}$ and $\tilde{\mW}^{(t)}_{\text{new}}=\mathbf{0}$, we substitute $\vp_{M^{(t-1)}:M^{(t)}}^{(t)}=\frac{\vx_{M^{(t-1)}:M^{(t)}}^{(t)}}{\sum_{l=M^{(t-1)}}^{M^{(t)}} (x_l^{(t)})^2}$ into \cref{eq:grohess_weight_dynamics_solved} and see that the contribution of the old weights to the weights learned for the new task $t$ vanishes.

\begin{equation}\label{eq:grohess_weight_dynamics_solved}
\mW^{(t)}_{\text{new}}
= \tilde{\mW}^{(t)}_{\text{new}} \left( \mI - \vx^{(t)}_{:,M^{(t-1)}:M^{(t)}}\frac{\vx^{(t),T}_{M^{(t-1)}:M^{(t)}}}{\sum_{l=M^{(t-1)}}^{M^{(t)}} (x_l^{(t)})^2} \right) + y^{(t)}\frac{\vx_{M^{(i-1)}:M^{(i)}}^{(t),T}}{\sum_{l=M^{(t-1)}}^{M^{(t)}} (x^{(t)}_{l})^2}
\end{equation}

Note that this simplifies to a simple least-squares estimator of $y^{(t)}$ in terms of the new hidden units of $\vx^{(t)}$.
Thus, the contribution of the old hidden units vanishes for the added dimensions, and the "new" parameters entirely absorb the new association between $\vx^{(t)}$ and $y^{(t)}$.
This concludes the proof for Corollary 1.

\subsubsection*{Implications}
In conclusion, this simplified but analytically tractable continual learning task demonstrates that \gls{noracl} strictly uses new hidden units to absorb incoming task information while preserving the representations of old units completely. 
Since the number of hidden units is fixed for \gls{ewc} regularization on a purely static network, over time the weights associated with any new task $i$ cannot be added to the old weights without overwriting important old task information.
Accordingly, the information about new tasks that can be encoded in the parameters decays proportionally to $t^{-1}$.
\gls{noracl} solves this by linearly increasing the hidden layer width as new tasks are introduced (see \cref{eq:size_scaling}).
Note that this linear scaling of the hidden layer width perfectly balances the inverse scaling of the weight decay in static networks.
Essentially, \gls{noracl} is able to avoid catastrophic forgetting as well as power-law forgetting by increasing its own capacity commensurate to the task demands, and then routing new gradients into the highly plastic, unconstrained new parameters.

\subsubsection{Transfer of Theory to ReLU MLP Experiments}\label{appendix:theory_transfer}

\begin{figure*}[h!]
    \centering
    \includegraphics[width=0.9\textwidth]{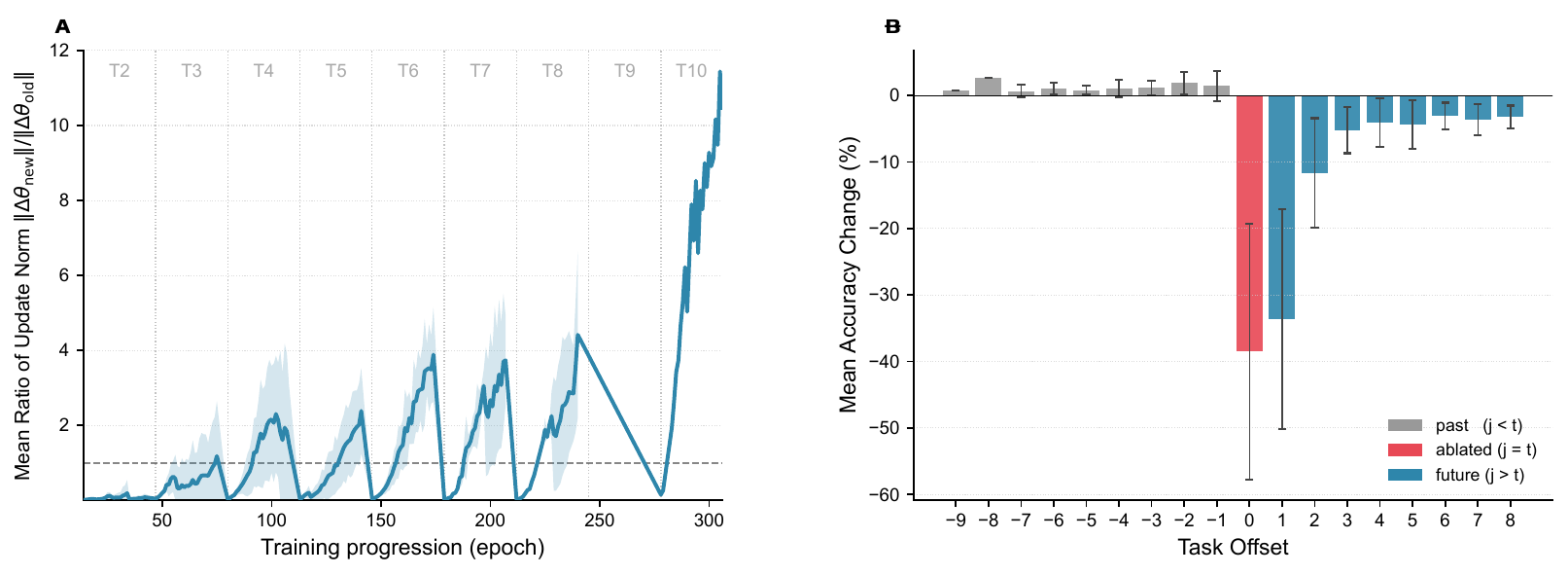}
    \caption{%
       \textbf{Old parameters are preserved and newly grown neurons absorb new task information.}
        (a) Mean ratio of update norms on new parameters versus old parameters over training. Error bars indicate standard deviation across 5 seeds for all epochs in which new neurons were grown. (b) Mean accuracy change across earlier $(j<t)$ and later $(j>t)$ tasks when ablating neurons grown during a task $t$ on the permuted MNIST sequence. Error bars indicate standard deviation across 5 seeds and all tasks.}
    \label{fig:app_theory}
\end{figure*}

Our simplified random feature setting is useful because it is analytically tractable. The resulting theory provides us with a set of hypotheses and a mechanistic intuition about how \gls{noracl} helps the network successfully navigate the stability-plasticity tradeoff during training. 
To empirically verify our theory in a more realistic setting, we investigate parameter update norms and neuron ablation impact in ReLU MLPs trained on permuted MNIST. Our main predictions from the theory (\cref{sec:theory}) are that old parameters are preserved relative to new parameters (Theorem 1) and that newly grown neurons absorb the bulk of new task information at the time they are added (Corollary 1).

\Cref{fig:app_theory}a plots the mean ratio of update norms on new parameters versus old parameters over training. Note that the ratio spikes shortly after adding new neurons, indicating that update norms on new parameters exceed those on old parameters (after enough time has passed for the fan-out zero initialization to be unlearnt and gradients to reach the new parameters). Thus, our first theoretical prediction qualitatively holds: relative to the updates performed on new parameters, old parameters are mostly preserved.

In \Cref{fig:app_theory}b, we evaluate the mean accuracy change across the different permuted MNIST tasks $j$ when ablating the set of neurons that were grown during the training of task $t$. This mean accuracy change is plotted as a function of the task offset $(j-t)$, allowing us to investigate the impact (after training on the entire sequence of tasks) of grown parameters on earlier tasks, later tasks, and the task they were grown for. Ablating neurons grown for task $t$ has a negligible, or even slightly positive, impact on the accuracy of earlier tasks, indicating slight but practically negligible retroactive interference. Later tasks rely far more on previously grown parameters (indicating forward transfer), however the ablation impact quickly decays for more distant future tasks. Finally, ablating grown parameters decreases performance the most on the task that they were grown for. Overall, this ablation study confirms our second theoretical prediction: newly grown capacity absorb the bulk of new task information within a short time of being added.

Overall, our theoretical model, despite its simplified setting, successfully explains how and where deep nonlinear networks trained with \gls{noracl} allocate plasticity when confronted with new task information. Transferring theoretical insights from simplified (e.g., shallow and linear) to realistic (e.g., deep and nonlinear) settings has been repeatedly shown to yield practical and actionable insights for understanding how deep neural networks learn. Our theoretical setting falls well within this tradition, combining the random pattern association task, used by \citet{kirkpatrick2017EWC} to analyse the original \gls{ewc}, with the random feature model, introduced by \citet{rahimi2007randomfeat}, to analyse how \gls{noracl} enables networks to balance stability and plasticity during continual learning.

\subsection{Additional results on interpretable architectures}\label{sec:app_interpretable}

\begin{figure*}[h!]
    \centering
    \includegraphics[width=0.45\textwidth]{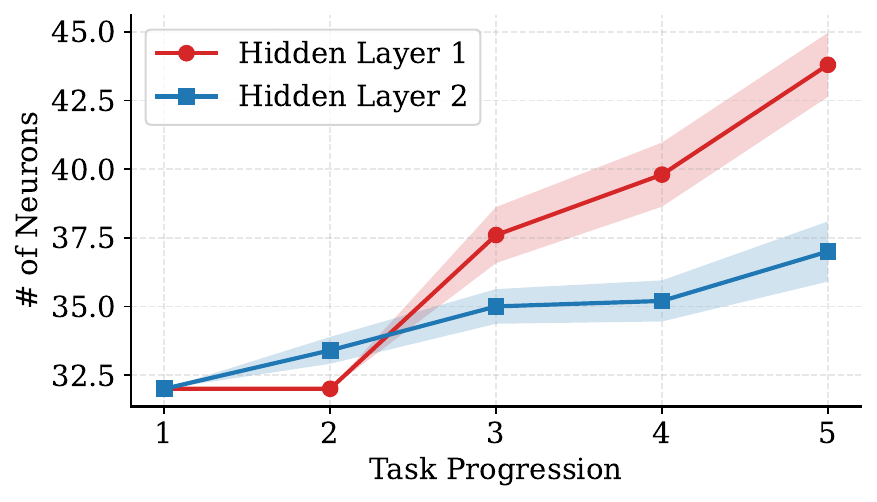}
    \caption{%
        \textbf{Layer-wise growth on Rotated MNIST.}
        Number of neurons per hidden layer as a function of task progression for the 2-layer \gls{noracl} model on Rotated MNIST. In contrast to the strongly asymmetric growth patterns observed on Permuted and Binary Split MNIST in \Cref{fig:F3}, both layers grow here, but layer 1 expands more than layer 2, consistent with rotations altering pose while preserving some low-level local structure.}
    \label{fig:app_rmnist}
\end{figure*}

\FloatBarrier

\subsection{Plasticity metrics}\label{sec:app_plasticity_diagnostics}

We derive two metrics for quantifying the plasticity state of a network using a per-parameter curvature information (like the diagonal Fisher Information Matrix). Both are derived directly from the per-parameter \emph{effective learning rate} induced by online \gls{ewc} \citep{schwarz2018progress} and, to our knowledge, have not been formalized as continual learning metrics in prior work. Recall from ~\Cref{sec:methods} that the \gls{ewc} penalty contributes a term $\tfrac{\lambda}{2}\,\tilde{F}_i (\theta_i - \theta_i^*)^2$ to the loss for parameter $i$, so the gradient step on $\theta_i$ is rescaled relative to the base learning rate $\eta$ by a factor that decreases as the accumulated Fisher $\tilde{F}_i$ grows. We define the \emph{effective learning rate} of parameter $i$ as
\begin{equation}
\eta^{\mathrm{eff}}_i \;=\; \frac{\eta}{1 + \lambda\,\tilde{F}_i},
\end{equation}
which captures, to first order, how much the optimizer can still move that parameter under the current \gls{ewc} constraint. When $\lambda\tilde{F}_i$ is small, $\eta^{\mathrm{eff}}_i \approx \eta$ (parameter is fully plastic), when $\lambda \tilde{F}_i \gg 1$, $\eta^{\mathrm{eff}}_i \to 0$ (i.e. parameter is frozen).

Building on this, we define two scalar diagnostics, the \textbf{locked fraction} of a layer $\ell$, with $|\ell|$ parameters, is the fraction of its parameters whose effective learning rate has fallen below 10\% of the base rate:
\begin{equation}
\mathrm{LockedFrac}_\ell \;=\; \frac{1}{|\ell|}\sum_{i \in \ell} \mathbb{1}\!\left[\, \eta^{\mathrm{eff}}_i < 0.1\,\eta \,\right].
\label{eq:locked_frac}
\end{equation}
A high locked fraction indicates that most of the layer has been frozen by accumulated Fisher and is no longer available for new learning. The \textbf{effective plastic parameter count} of the network is the sum of normalized effective learning rates across \emph{all} parameters:
\begin{equation}
N^{\mathrm{eff}}_{\mathrm{plastic}} \;=\; \sum_{\ell}\sum_{i \in \ell} \frac{\eta^{\mathrm{eff}}_i}{\eta} \;=\; \sum_{\ell}\sum_{i \in \ell}\frac{1}{1 + \lambda\,\tilde{F}_i}.
\label{eq:epp}
\end{equation}
Intuitively, $N^{\mathrm{eff}}_{\mathrm{plastic}}$ is the \emph{effective number of fully-plastic parameters} the network still has: a fully unconstrained parameter contributes 1, a fully frozen parameter contributes 0, and partially constrained parameters contribute fractionally. Unlike the locked fraction, this metric is threshold-free and gives a continuous measure of how much usable plastic capacity the optimizer has access to at any point in training.

\subsection{Extended long-horizon stress test: 100-task Permuted MNIST}
\label{sec:app_long_horizon_100}

\begin{figure*}[t]
    \centering
    \includegraphics[width=\textwidth]{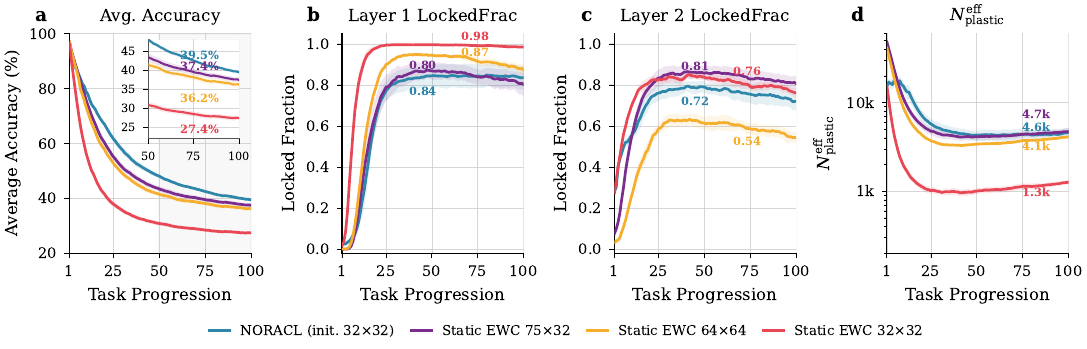}
    \caption{%
    \textbf{100-task Permuted MNIST: stress test.}
    \textbf{(a)}~Average accuracy across all seen tasks. 
    Inset: zoom into tasks 50--100.
    \textbf{(b, c)}~Per-layer locked fraction 
    ($\mathrm{LockedFrac}_\ell$, Eq.~\ref{eq:locked_frac}).
    \textbf{(d)}~Effective plastic parameter count 
    $N^{\mathrm{eff}}_{\mathrm{plastic}}$ (Eq.~\ref{eq:epp}, 
    log scale)
    }
    \label{fig:app_F4_100}
\end{figure*}

\Cref{fig:app_F4_100} reports four metrics over training for the static and \gls{noracl} conditions: (a) average accuracy across all seen tasks, (b) hidden layer-1 locked fraction $\mathrm{LockedFrac}_1$, (c) hidden layer-2 locked fraction $\mathrm{LockedFrac}_2$, and (d) effective plastic parameter count $N^{\mathrm{eff}}_{\mathrm{plastic}}$. Across all four panels, a consistent picture emerges: static networks of any size undergo monotonic plasticity exhaustion, which larger initial capacity can delay but not remove, whereas \gls{noracl} counteracts this trend through incremental growth. 

In panel~(a), all three static baselines degrade over time, with smaller networks collapsing faster. \gls{noracl} achieves the highest final average accuracy, including relative to the architecture-matched $75\times32$ control whose final layer widths closely match its own. This comparison is particularly informative: the $75\times32$ model benefits early from having its full capacity available from the start, but is later overtaken as its parameters accumulate Fisher importance. \gls{noracl}'s advantage therefore lies not in ending with a larger architecture, but in the \emph{temporal distribution of capacity}: new neurons enter the network with $\tilde{F}=0$ and are fully plastic precisely when additional capacity is needed. 

Panels~(b) and~(c) show that all static networks converge toward high locked fractions, with smaller networks saturating earlier and more severely. \gls{noracl} maintains the lowest locked fraction in the first hidden layer despite starting from the smallest initial width, and remains competitive or lower in the second layer as well. Because \gls{noracl}'s layer widths are themselves increasing over time, a relatively flat locked fraction implies a replenishment dynamic: fresh plastic parameters are being added at roughly the rate existing ones become constrained. No static architecture can exhibit this behavior, since its parameter set is fixed from the start. 

Panel~(d) helps explain why \gls{noracl} can outperform static networks even when their final effective plastic counts are similar. By the end of the stream, \gls{noracl} and the $75\times32$ baseline retain comparable values of $N^{\mathrm{eff}}_{\mathrm{plastic}}$, yet \gls{noracl} attains higher accuracy. A natural interpretation is that the composition of this plasticity differs. In the static $75\times32$ network, the surviving plastic parameters are residual survivors of a long consolidation process and therefore carry accumulated Fisher from many tasks. In \gls{noracl}, part of the plasticity budget comes from neurons that entered with $\tilde{F}=0$ later in the stream and remained fully plastic until subsequent consolidation. Thus, not all plastic parameters are equally plastic: fresh capacity appears more useful than residual capacity, which helps explain the remaining accuracy gap. 

A minor limitation here is that the locked-fraction diagnostic depends on a cutoff on $\eta^{\mathrm{eff}}/\eta$ (we use $0.1$). While we have verified that the qualitative gap between \gls{noracl} and the static baseline is robust across thresholds, it should be read as a diagnostic trend rather than a precise measure. The effective-plastic-parameter count $N^{\mathrm{eff}}_{\mathrm{plastic}}$, on the other hand, is threshold-free by construction and could be read as the primary plasticity diagnostic, with $\mathrm{LockedFrac}_\ell$ serving as an interpretable per-layer breakdown.

\subsection{Ablation Studies}\label{sec:app_ablations}

\gls{noracl}'s growth mechanism must answer three questions at each expansion event: \emph{when} to grow (is the current architecture insufficient?), \emph{where} to grow (which layer needs additional capacity?), and \emph{how} to grow (how should the new neurons be initialized?). The first two questions are addressed jointly by the growth trigger (\Cref{sec:method_trigger}), which combines the spectral effective dimension (\gls{ed}) signal with the Fisher saturation ($F_{\mathrm{sat}}$) signal; the third is addressed by the QR-based orthogonal initialization (\Cref{sec:method_growth}). We ablate each component in turn, reporting results on all three benchmarks in ~\Cref{tab:ablations}, in addition to comparing to alternative growth criteria used in prior work on expansion-based continual learning.

\begin{table*}[t]
\centering
\begin{threeparttable}
\caption{Ablation study of initialization strategies and growth triggers across continual learning benchmarks.}
\label{tab:ablations}
\renewcommand{\arraystretch}{1.15}
\setlength{\tabcolsep}{5pt}

\resizebox{\textwidth}{!}{%
    \begin{minipage}{\textwidth} %
    \begin{tabular}{llcccccc}
    \toprule
    \multirow{2}{*}{\textbf{Ablation}} & \multirow{2}{*}{\textbf{Variant}} 
    & \multicolumn{2}{c}{\textbf{Permuted MNIST}} 
    & \multicolumn{2}{c}{\textbf{Rotated MNIST}} 
    & \multicolumn{2}{c}{\textbf{Binary Split MNIST}} \\
    \cmidrule(lr){3-4} \cmidrule(lr){5-6} \cmidrule(lr){7-8}
    & & \textbf{Params} & \textbf{Acc. (\%)} 
      & \textbf{Params} & \textbf{Acc. (\%)} 
      & \textbf{Params} & \textbf{Acc. (\%)} \\
    \midrule
    \multirow{1}{*}{\textbf{Full model}}
    & -        & 49.2k & 79.4 $\pm$ 0.7 & 42.9k & \textbf{74.9 $\pm$ 1.7} & 23.3k & \textbf{73.9 $\pm$ 2.5} \\
    \midrule
    \multirow{5}{*}{\textbf{Initializations}}
    & Random init    & 47.4k $\pm$ 2.6 & 65.9 $\pm$ 2.5 & 36.0k $\pm$ 1.3 & 67.6 $\pm$ 3.6 & 40.9k $\pm$ 9.2 & 71.5 $\pm$ 4.8 \\
    & He init        & 64.3k $\pm$ 2.4 & 76.1 $\pm$ 0.8 & 38.0k $\pm$ 1.3 & 71.0 $\pm$ 1.7 & 34.0k $\pm$ 6.6 & 71.1 $\pm$ 3.8 \\
    & Xavier init    & 64.2k $\pm$ 5.8 & 76.5 $\pm$ 1.2 & 38.6k $\pm$ 1.7 & 70.0 $\pm$ 2.6 & 35.7k $\pm$ 6.3 & 71.5 $\pm$ 3.2 \\
    & Null-space init& 58.0k $\pm$ 2.8 & 78.1 $\pm$ 0.9 & 41.8k $\pm$ 2.8 & 72.6 $\pm$ 3.6 & N/A\tnote{a} & N/A\tnote{a} \\
    & Zero init      & 39.3k $\pm$ 3.3 & 63.0 $\pm$ 3.7 & 35.4k $\pm$ 0.8 & 67.7 $\pm$ 2.4 & 31.3k $\pm$ 3.4 & 70.7 $\pm$ 3.8 \\
    \midrule

    \multirow{5}{*}{\textbf{Growth Triggers}}
    & ED only                         & 356.8k $\pm$ 7.2 & \textbf{81.2 $\pm$ 0.4} & 114k $\pm$ 6.2 & 72.8 $\pm$ 2.4 & 78.2k $\pm$ 20.6 & 70.7 $\pm$ 2.9 \\
    & $F_{\mathrm{sat}}$ only         & 46.3k $\pm$ 2.8 & 74.2 $\pm$ 1.1 & 37.2k $\pm$ 1.3 & 71.5 $\pm$ 2.6 & 30.7k $\pm$ 4.3 & 70.0 $\pm$ 2.8 \\
    & Fixed growth / task             & 75.6k & 53.8 $\pm$ 0.6 & 47.6k & 62.3 $\pm$ 1.2 & 29.8k & 59.0 $\pm$ 0.3 \\
    & Scheduled growth $\{0,5,15,25\}$ & 93.3k & 77.5 $\pm$ 0.2 & 54.9k & 71.4 $\pm$ 2.7 & 36.8k & 61.4 $\pm$ 0.5 \\
    & Loss-based                     & 42.2k $\pm$ 1.5 & 74.4 $\pm$ 2.1 & 35.5k $\pm$ 1.8 & 70.6 $\pm$ 2.3 & 16k $\pm$ 2 & 70.9 $\pm$ 3.4 \\
    \bottomrule

    \end{tabular}

    \begin{tablenotes}
    \footnotesize
    \item[a] Null-space initialization was infeasible in 4/5 seeds on Binary Split MNIST because the available null-space dimension was exhausted during growth. Results are therefore not reported as a 5-seed mean $\pm$ std.
    \end{tablenotes}
    \end{minipage}
}
\end{threeparttable}
\end{table*}

\subsubsection{Growth Triggers: When and Where to Grow}\label{sec:app_ablations_growth_trig}

The joint \gls{ed}~$\wedge$~$F_{\mathrm{sat}}$ trigger is designed so that each signal compensates for the other's failure mode. The \gls{ed} signal detects representational saturation (all neurons producing linearly independent features), but it can fire spuriously due to transient batch effects or optimization dynamics unrelated to genuine capacity needs. The $F_{\mathrm{sat}}$ signal confirms that the layer's parameters are genuinely being relied upon by the current task, but on its own it cannot determine \emph{which} layer needs expansion. \Cref{tab:ablations} confirms that both signals are necessary and that neither alone is sufficient.

\paragraph{ED only:} Removing the Fisher saturation signal causes dramatic over-growth. On Permuted MNIST, the network inflates to $356.8$k parameters, over $7\times$ the size of the full \gls{noracl} model ($49.2$k), for a marginal accuracy gain of $1.8\%$ ($81.2\%$ vs.\ $79.4\%$). On Rotated MNIST the over-growth is similarly severe ($114$k vs.\ $42.9$k) with \emph{lower} accuracy ($72.8\%$ vs.\ $74.9\%$), and on Binary Split MNIST the \gls{ed}-only network grows to $78.2$k with high variance ($\pm 20.6$k) and again underperforms the full model ($70.7\%$ vs.\ $73.9\%$). The pattern is clear, without the Fisher saturation signal the \gls{ed} signal fires on transient saturation events that do not reflect genuine capacity needs, producing networks that are wastefully large and, on benchmarks with shared task structure, actually \emph{less} accurate because the excess neurons dilute the gradient signal and complicate consolidation. Note that using only \gls{ed} as the growth signal is equivalent to deploying NORTH* \citep{maile2022and} per task while simultaneously using \gls{ewc} regularization. Thus, this ablation case serves as a natural expansion-based baseline to compare \gls{noracl} to.

\paragraph{$F_{\mathrm{sat}}$ only:} Removing the \gls{ed} signal produces the opposite failure mode: the network grows conservatively but insufficiently. On Permuted MNIST, the $F_{\mathrm{sat}}$-only model reaches only $74.2\%$ at $46.3$k parameters, compared to $79.4\%$ at $49.2$k for the full model, a $5.2\%$ accuracy deficit for a comparable parameter budget. The reason is that $F_{\mathrm{sat}}$ alone cannot determine \emph{which} layer is the bottleneck: Fisher importance accumulates in all layers simultaneously as tasks progress, so the signal fires indiscriminately rather than targeting the layer whose representational capacity is actually saturated. The result is growth that is poorly allocated across layers. On Rotated and Binary Split MNIST the pattern is consistent: $F_{\mathrm{sat}}$-only underperforms the full model by $3.5\%$ and $3.9\%$, respectively.

\paragraph{Heuristic triggers:} We additionally compare against two heuristic growth strategies that do not use either principled signal. \emph{Fixed growth per task} adds a constant number of neurons to every layer at the start of each task. This produces severe accuracy degradation across all benchmarks ($53.8\%$ on Permuted MNIST, $62.3\%$ on Rotated, $59.0\%$ on Binary Split), confirming that undirected growth is actively harmful. The network wastes capacity on layers that do not need it and disrupts learned representations through poorly timed expansion. \emph{Scheduled growth} adds neurons only at predetermined task indices $\{0, 5, 15, 25\}$. This performs better than fixed growth ($77.5\%$ on Permuted) because growth events are less frequent and concentrated early, but it still underperforms the full \gls{noracl} model on all three benchmarks and, crucially, requires the practitioner to choose a schedule which reintroduces the kind of oracle knowledge that \gls{noracl} is designed to eliminate. 

\paragraph{Loss-based trigger:} We also compare against a loss-based growth trigger that expands the network when validation performance on the current task plateaus over a short epoch window. This type of heuristic has precedent in earlier expansion-based continual learning methods such as \cite{yoon2017lifelong}, where growth is triggered after selective retraining encounters a loss plateau. However, as discussed in \Cref{sec:related_work}, such criteria conflate multiple causes such as genuine representational saturation or temporary optimization slowdown. In contrast, \gls{noracl}'s joint \gls{ed}~$\wedge$~$F_{\mathrm{sat}}$ trigger is designed to detect a more specific failure mode, namely that a particular layer is simultaneously saturated in its representation and too constrained to be repurposed.

\subsubsection{Initialization: How to Grow}\label{sec:app_ablations_init}

When new neurons are added, their fan-in weights must be initialized in a way that (i)~does not disrupt existing representations (zero fan-out ensures this by construction, \Cref{sec:method_growth}), and (ii)~gives the new neurons diverse, non-redundant input directions so that they can quickly specialize for the arriving task. We compare QR-based orthogonal initialization \citep{saxe2013exact} against five alternatives: standard random normal initialization, He initialization~\citep{he2015delving}, Xavier initialization~\citep{glorot2010understanding}, null-space initialization and zero initialization.

\paragraph{QR orthogonal initialization} achieves the highest accuracy on Rotated MNIST ($74.9\%$) and Binary Split MNIST ($73.9\%$), and is competitive on Permuted MNIST ($79.4\%$), while consistently producing the most compact architectures. The key advantage is that mutually orthogonal fan-in vectors ensure that new neurons begin from maximally diverse input directions, avoiding redundancy and enabling rapid differentiation during training. This is particularly important in continual learning, where the training window per task is short and newly added neurons must specialize quickly before consolidation begins to constrain them.

\paragraph{He and Xavier initialization} both achieve reasonable accuracy on Permuted MNIST ($76.1\%$ and $76.5\%$, respectively) but produce substantially larger networks ($64.3$k and $64.2$k vs.\ $49.2$k for QR). The explanation is that these methods draw fan-in vectors independently from a distribution calibrated for variance preservation, but do not enforce orthogonality between them. As a result, some new neurons start from near-redundant directions and fail to differentiate within the training window, triggering the growth signal again at the next evaluation and producing unnecessary subsequent expansion. On Rotated and Binary Split MNIST, the accuracy gap widens ($4$-$5\%$ below QR) and the variance across seeds increases, indicating less stable learning dynamics.

\paragraph{Random initialization} performs worst among the non-trivial methods ($65.9\%$ on Permuted), with both lower accuracy and higher cross-seed variance. Without scale calibration, the magnitude of the new fan-in weights is poorly matched to the existing network, leading to gradient-scale mismatches that slow integration of the new neurons.

\paragraph{Null-space initialization} where newly added neurons are initialized in directions orthogonal to the span of the existing incoming weights of the grown layer, performs reasonably well on Permuted MNIST ($78.1 \pm 0.9\%$) and Rotated MNIST ($72.6 \pm 3.6\%$), placing it closer to QR initialization. Conceptually, this is an appealing alternative to QR initialization, however, it is not guaranteed to remain feasible throughout training. Its availability depends on the remaining dimension of the layer's null space, which can shrink or vanish as the layer grows and approaches full rank. This limitation becomes visible on Binary Split MNIST, where the hidden layers remain relatively small and the available null-space dimension was exhausted in $4/5$ seeds, preventing a comparable 5-seed estimate from being reported. We therefore view null-space initialization as an informative ablation that supports the broader importance of \emph{novel} fan-in directions, but also as evidence that QR is the more robust practical choice. 

\paragraph{Zero initialization} is a useful control. Setting both fan-in and fan-out weights to zero makes growth trivially function-preserving, but the new neurons are permanently inert, i.e. they receive no gradient signal (because their activations are identically zero) and therefore contributes less to task learning. This confirms that the accuracy gains of \gls{noracl} over static baselines are indeed attributable to the learned contributions of the added neurons, not merely to the act of adding parameters.

These ablation results confirm that all three components of \gls{noracl}'s growth mechanism, the \gls{ed} signal, the Fisher confirmation, and the QR initialization, contribute meaningfully to its performance. The \gls{ed} signal provides layer-specific capacity monitoring, the Fisher signal prevents spurious over-growth and orthogonal initialization ensures that newly added neurons integrate efficiently within the short per-task training window. Removing or replacing any one of these components degrades either accuracy, parameter efficiency, or both.

\subsection{Computational Cost}
\label{sec:app_compute_cost}

We report wall-clock overhead of NORACL relative to the EWC backbone on Permuted MNIST, which is the worst case among our benchmarks as it triggers the most growth events (12 over 10 tasks), and any monitoring overhead scales with task count. On the more compact streams (Rotated and Binary Split MNIST), overhead is correspondingly lower. All measurements use a 2-layer MLP with initial hidden width $32{\times}32$, batch size $256$, $30$ epochs per task ($10$ for task 1), and are averaged over 5 seeds on a single \texttt{RTX4090} GPU with no other jobs sharing the device. Validation and CL evaluation costs are identical across methods and are excluded from the overhead calculation.

\begin{table}[h]
\centering
\caption{Wall-clock breakdown on 10-task Permuted MNIST (2-layer MLP,
single GPU). Shared components are incurred by both NORACL and a plain-EWC
baseline; NORACL-only components are the additional cost of the growth
mechanism. All percentages are of total wall-clock.}
\label{tab:compute}
\setlength{\tabcolsep}{8pt}
\begin{tabular}{lrrl}
\toprule
Component & Time (s) & \% of total & Attribution \\
\midrule
SGD training        & 481.8 & 72.5 & shared \\
Validation eval     & 133.9 & 20.1 & shared \\
EWC consolidation   &   2.0 &  0.3 & shared \\
CL evaluation       &  12.2 &  1.8 & shared \\
\midrule
ED / SVD monitoring &   6.1 &  0.9 & NORACL only \\
Fisher growth-check &  13.8 &  2.1 & NORACL only \\
Neurogenesis        &  15.3 &  2.3 & NORACL only \\
\midrule
\textbf{NORACL overhead} & \textbf{35.1} & \textbf{5.3}$^{\dagger}$ & --- \\
Total wall-clock    & 664.9 & 100.0 & --- \\
\bottomrule
\end{tabular}\\[0.3em]
{\footnotesize $^{\dagger}$ The growth mechanism takes 35.1\,s, which is 5.3\% of NORACL's total wall-clock (664.9\,s) and 5.6\% of a plain-EWC baseline's runtime (629.9\,s, the shared components only). On single GPU, 12 growth events over the stream.}
\end{table}

The largest NORACL-specific cost is the per-epoch Fisher growth-check ($2.1\%$ of wall-clock), followed by the spectral monitoring ($0.9\%$). This is because the growth-check requires a forward and backward pass on a few mini-batches to compute per-parameter gradients for $F^{(\text{curr})}_l$, whereas the spectral signal requires only a single SVD on the layer's activation matrix $H_l \in \mathbb{R}^{M_l \times n}$. At the layer widths considered here ($M_l \leq 75$), the SVD is well below one second per epoch.

Across the 12 growth events on Permuted MNIST, neurogenesis itself (QR initialization of fan-in weights, zero-padding of fan-out weights and Fisher/anchor buffers, optimizer state reinitialization) accounts for $15.3$~s in aggregate, or approximately $1.27$~s per event. This cost is amortized over subsequent training and does not scale with task count beyond the number of events themselves. On streams with fewer growth events (e.g.\ Binary Split MNIST), this contribution is correspondingly smaller.

The two NORACL-specific monitoring costs scale differently with model size. The per-epoch Fisher check scales as $\mathcal{O}(P \cdot B \cdot n)$ where $P$ is the layer's parameter count, $B = 5$ is the number of mini-batches used for the within-epoch estimate, and $n = 256$ is the batch size, for a total of 1,280 samples per check. It is dominated by standard forward and backward passes and inherits any hardware optimization available to those. The spectral check scales as $\mathcal{O}(\min(M_l, n)^2 \cdot \max(M_l, n))$ per layer per epoch. This is the standard cost of a SVD on the $M_l \times n$ activation matrix, set by the smaller dimension (which bounds the spectrum's size) squared times the larger dimension. At the widths in our experiments, both costs are sub-second per epoch. For substantially wider layers ($M_l \gg 10^3$), the SVD becomes the asymptotically dominant component and can be replaced by randomized SVD~\citep{halko2011finding}, which reduces its cost to $\mathcal{O}(M_l \cdot n \cdot k)$ for a target rank $k$. We did not need this optimization for the present results.

NORACL's growth mechanism is a small additive cost on top of the EWC backbone rather than a fundamentally more expensive procedure. The overhead is dominated by the Fisher growth-check, which reuses the gradient infrastructure already present for EWC consolidation. The spectral check is cheap in absolute terms at the scales considered, with a well-understood asymptotic replacement available for wider networks.

\subsection{Sensitivity to growth-trigger thresholds}
\label{sec:app_sensitivity}

NORACL's growth mechanism is governed by three thresholds, the ED discount $\gamma$, which controls how close a layer must be to representational saturation before growth fires, the Fisher percentile $p$, which controls how large a fraction of a layer's parameters must be plasticity-saturated and the SVD threshold $\varepsilon$, which sets the cutoff below which a singular value is not counted toward the effective dimension. To assess how robust NORACL is to these choices, we vary each threshold one at a time around the default configuration used throughout the main experiments, holding all other hyperparameters fixed. All runs use Permuted MNIST with the 2-layer model and five seeds, we report final average accuracy and total parameter count.

\begin{table}[h]
\centering
\caption{Sensitivity of NORACL to growth-trigger thresholds. We vary each
hyperparameter one at a time around the default configuration and report
final average accuracy and parameter count on Permuted MNIST over 5 seeds.
The default row matches the 2L NORACL entry in Table~\ref{tab:NORACL_static_detailed}.}
\label{tab:sensitivity}
\setlength{\tabcolsep}{6pt}
\begin{tabular}{llrr}
\toprule
\multicolumn{2}{l}{Variant} & Params & Acc (\%) \\
\midrule
\multicolumn{2}{l}{Default$^{\dagger}$\ \ ($\gamma=0.9,\ p=25,\ \varepsilon=0.05$)}
   & 49.2k $\pm$ 3.2 & 79.4 $\pm$ 0.7 \\
\midrule
\multicolumn{4}{l}{\textit{ED discount $\gamma$ \ (representational-saturation discount factor)}} \\
 & $\gamma = 0.80$ & 67.7k $\pm$ 7.4 & 81.2 $\pm$ 1.2 \\
 & $\gamma = 0.85$ & 64.1k $\pm$ 3.3 & 80.3 $\pm$ 1.6 \\
 & $\gamma = 0.95$ & 54.0k $\pm$ 3.7 & 79.4 $\pm$ 1.4 \\
\midrule
\multicolumn{4}{l}{\textit{Fisher percentile $p$ \ (plasticity-saturation threshold)}} \\
 & $p = 10$ & 34.1k $\pm$ 3.6 & 68.0 $\pm$ 5.0 \\
 & $p = 15$ & 41.6k $\pm$ 3.9 & 73.9 $\pm$ 2.8 \\
 & $p = 20$ & 53.7k $\pm$ 2.7 & 77.4 $\pm$ 1.2 \\
 & $p = 30$ & 70.3k $\pm$ 1.7 & 79.0 $\pm$ 1.8 \\
\midrule
\multicolumn{4}{l}{\textit{SVD threshold $\varepsilon$ \ (effective-dimension counting)}} \\
 & $\varepsilon = 0.01$ & 61.6k $\pm$ 3.3 & 80.9 $\pm$ 1.6 \\
 & $\varepsilon = 0.10$ & 58.4k $\pm$ 2.7 & 79.8 $\pm$ 1.2 \\
\bottomrule
\end{tabular}\\[0.3em]
{\footnotesize $^{\dagger}$ Default row reproduces the 2L NORACL Permuted MNIST
result from Table~\ref{tab:NORACL_static_detailed}, included for reference.}
\end{table}

\textbf{ED discount $\gamma$.} Accuracy and parameter count both vary smoothly and monotonically with $\gamma$. More aggressive discounting (smaller $\gamma$, growth fires earlier relative to full saturation) yields larger networks and marginally higher accuracy. Across the full range, accuracy varies by under $2$ percentage points and no setting fails. The smooth dependence of parameter count on $\gamma$ is exactly what the mechanism predicts, i.e. $\gamma$ sets how proactively growth responds to representational pressure, so smaller values admit more growth events.

\textbf{Fisher percentile $p$.}
The Fisher percentile is the most consequential of the three thresholds. Lowering $p$ makes the plasticity-saturation condition stricter which suppresses growth and produces undercapacity networks. At $p=10$ accuracy falls to $68.0\%$ with high seed variance ($\pm 5.0\%$), reflecting that growth has become so rare that whether it occurs at all is seed-dependent. As $p$ increases toward the default, accuracy recovers and the parameter budget grows accordingly. Crucially, accuracy \emph{saturates} near the default, moving from $p=25$ to $p=30$ enlarges the network by roughly $40\%$ ($49.2$k to $70.3$k parameters) for no accuracy gain ($79.4\%$ to $79.0\%$, within seed noise). The default therefore sits at the accuracy-capacity trade-off, in the parameter-efficient regime the method targets.

\textbf{SVD threshold $\varepsilon$.}
NORACL is essentially insensitive to $\varepsilon$. Both $\varepsilon = 0.01$ and $\varepsilon = 0.10$ fall within $1.5$ percentage points of the default and within seed variance, and the parameter counts differ only modestly. The singular-value spectrum of the layer activations is sufficiently well-separated that the precise cutoff has little effect on the count. We therefore treat $\varepsilon$ as a fixed numerical constant rather than a tuned hyperparameter.

Across all three thresholds, NORACL degrades gracefully and exhibits no catastrophic failure within reasonable ranges. The ED discount $\gamma$ and SVD threshold $\varepsilon$ are robust, with accuracy varying by under $2$ points across the swept values. The Fisher percentile $p$ is the most sensitive, but the default value sits at the accuracy-capacity trade-off, which supports our use of a single fixed configuration across benchmarks (Section~\ref{sec:experiments}).

\end{document}